\documentclass[final]{cvpr}

\usepackage{times}
\usepackage{epsfig}
\usepackage{graphicx}
\usepackage{amsmath}
\usepackage{amssymb}
\usepackage[table]{xcolor}

\usepackage[pagebackref,breaklinks,colorlinks]{hyperref}

\def\etal{\emph{et al}.}
\def\cvprPaperID{5} 
\def\confName{CVPR Workshop}
\def\confYear{2021}
\graphicspath{{./figs/}{./figs/result/}}

\begin{document}

\title{DynaDog+T: A Parametric Animal Model for Synthetic Canine Image Generation.}

\author{Jake Deane\textsuperscript{1}  Sin\'{e}ad Kearney\textsuperscript{1}      Kwang In Kim\textsuperscript{2} Darren Cosker\textsuperscript{1}\\
CAMERA University of Bath\textsuperscript{1} Ulsan National Institute of Science and Technology\textsuperscript{2}\\
{\tt\small \textsuperscript{1}(j.deane, d.p.cosker)@bath.ac.uk \textsuperscript{2}kimki@unist.ac.kr}
}

\maketitle

\begin{abstract}
Synthetic data is becoming increasingly common for training computer vision models for a variety of tasks. Notably, such data has been applied in tasks related to humans such as 3D pose estimation where data is either difficult to create or obtain in realistic settings. Comparatively, there has been less work into synthetic animal data and it's uses for training models. Consequently, we introduce a parametric canine model, DynaDog+T, for generating synthetic canine images and data which we use for  a  common  computer  vision  task, binary segmentation, which would otherwise be difficult due to the lack of available data.

\end{abstract}

\section{Introduction}


The use of synthetic data for training deep neural networks for computer vision tasks is becoming more common \cite{nikolenko_synthetic_2019, varol_synthetic_2020, dunn_deepsynth_2019, li_synthetic_2021, sanchez_multi-task_2019, tremblay_training_2018} particularly for problems where there is either insufficient or non-existent labelled data. One notable example of this is animal based tasks such as animal pose estimation. While there has been a recent uptake in the amount of work related to animals\cite{cao_cross-domain_2019, khan_animalweb_2019, joska_acinoset_2021, vidal_perspectives_2021, parham_animal_2018, zuffi_3d_2016, zuf_lions_2018, zuf_three-d_2019}, compared to humans\cite{he_mask_2018, kanazawa_end--end_2017, zheng_deep_2021, ansari_human_2021, long_pp-yolo_2020} there is significantly less work due to the difficulty of obtaining sufficient labelled data. Therefore, we present a novel parametric canine model, DynaDog+T (Dynamic Dog plus Texture) which can be used to generate synthetic RGB canine images with paired labelled data: 2D and 3D joint positions, part segmentation maps and binary segmentation (a.k.a silhouette) maps. We also evaluate this data for a common computer vision task where there is real data available; segmentation. We do this to  answer an important question about our synthetic data: Is our synthetic data useful for training models where real data is either insufficient or non-existent.

\section{Related Work}

\textbf{Synthetic Data for Training:}
 In machine learning, particularly for deep learning, large amounts of data are required to train models. For some problems especially those in computer vision, such as those involving animals, as noted above, this is a problem as the acquisition of large scale labelled data is expensive and time consuming to produce not to mention it is difficult to capture all the variability present within the domain. Synthetic data allows one to circumvent this issue, via creating high quality artificial data with associated data points to either augment existing data or to create new datasets for training, where domain adaption is used to close the domain gap with real data\cite{tremblay_training_2018, sankaranarayanan_learning_2018, barth_improved_2018, li_synthetic_2021, barth_improved_2018, tremblay_training_2018}. This is hardly a problem unique to computer vision however, synthetic data is used in such areas as neural programming, bioinformatics and natural language processing \cite{nikolenko_synthetic_2019, hittmeir_utility_2019, mayer_what_2018}. In relation to this work perhaps the most important works are those of Varol \etal\cite{varol_learning_2017, varol_synthetic_2020}, where the  SMPL model\cite{loper_smpl:_2015} is used, in conjunction with motion capture data to create synthetic RGB images of humans (with pared data points such as part segmentation maps), Mu \etal\cite{mu_learning_2020}, where domain adaption techniques are used to close the gap between real and synthetic CAD model based images for animal pose estimation, and the work of Li and Lee \cite{li_synthetic_2021} who use domain adaption for animal pose estimation, building upon the work of Mu \etal \cite{mu_learning_2020}.


\textbf{Parametric Body Models:} Parametric modeling of 3D body shape is widely used to create realistic human bodies. Arguably the most famous and impact-full work in this area is SMPL\cite{loper_smpl:_2015} and related methods used to capture its parameters from images \cite{madadi_smplr:_2018, bogo_keep_2016, kanazawa_end--end_2017}. Two notable related works are Pavlakos \etal \cite{pavlakos_expressive_2019} and Osman \etal \cite{vedaldi_star_2020} which extend SMPL to enable greater variation and generalisation ability.  The work of Santesteban \etal \cite{santesteban_softsmpl_2020} extends SMPL via modelling soft tissue dynamics as a function of body shape and motion. SMPL has a for all intents and purposes, equivalent animal piece in Skinned Multi-Animal Linear (SMAL) model \cite{zuffi_3d_2016} which like its human counterpart has its own extensions and has seen use in capturing model parameters from images \cite{zuf_lions_2018, zuf_three-d_2019}. The most important piece of work as it pertains to our own, is that of Kearney \etal \cite{kearney_rgbd-dog_2020} who proposed a canine shape model trained using motion capture data which we utilised in the creation of our own model.

\textbf{Animals in Computer Vision:} As mentioned above, compared to human subjects, there is comparatively little real data available for animal domain problems within computer vision. Animal species form parts of existing computer vision datasets notably those related to segmentation and classification tasks \cite{lin_microsoft_2014, krasin_openimages_2017, kuznetsova_open_2018, khosla_novel_2011}. Similarly, datasets and methods have been created for part segmentation for common animal classes such as dogs and cows \cite{haggag_semantic_2016, wang_semantic_2015, chen_detect_2014}. In recent years, there has been significant progress in the field of 2D joint estimation for animals due to the introduction of numerous datasets and models \cite{labuguen_macaquepose_nodate, bala_openmonkeystudio_2020, khan_animalweb_2019, joska_acinoset_2021, biggs_who_2020, biggs_creatures_2018, cao_cross-domain_2019} and software that enables key-point labelling \cite{mathis_markerless_2018, mathis_pretraining_2019, mathis_pretraining_nodate} to create new labelled data. 


\section{DynaDog+T: Generative 3D Model for Canine Shape and Texture}

\begin{figure}
    \centering
    \includegraphics[width=\linewidth]{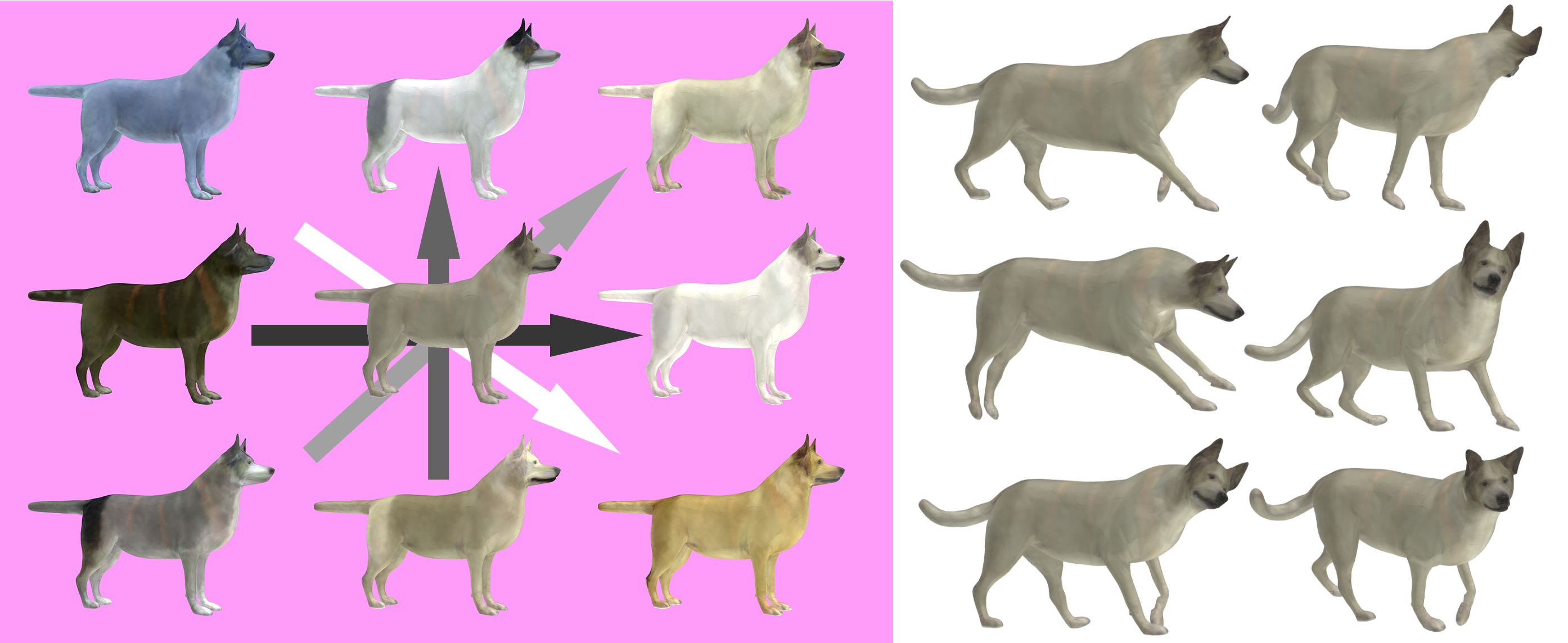}
    \caption{Left: First 4 principal components of the texture PCA model, displayed on a generic dog mesh. The mean texture is in the middle, darkest arrow denotes the first component, lightest arrow denotes the fourth component. Each component shows $\pm2$ standard deviations. 
    Right: Motion from Kearney \etal \cite{kearney_rgbd-dog_2020} applied to the dog with mean texture.
    }
    \label{fig:textureVariaton}
\end{figure}

We introduce a novel generative parametric model for canines, from which dataset instances of arbitrary size and variability can be generated; thus allowing the development of canine related computer vision models for a variety of tasks using this synthetic data. Our generative model, DynaDog+T (Dynamic Dog with Texture), improves and modifies the canine shape model proposed by Kearney \etal \cite{kearney_rgbd-dog_2020} in several key ways. The first difference is that we use a refined set of meshes to create the underlying PCA model used to generate shape -- where each 3D mesh has been increased in detail and hence overall quality by an artist. The second difference is that we introduce a new PCA model created from 12 UV texture maps as represented by Kato \etal \cite{kato_neural_2017}. These textures were produced through photogrammetry scans and scanned toy animals and were manually cleaned. All textures are shown in the supplementary material and the first four components of the Texture PCA shape space can be seen in the left of Figure \ref{fig:textureVariaton}.

Each texture is transformed into a matrix $T_i \in \mathbb{R}^{f\times d \times d \times d \times 3}$, where $f$ is the number of faces on the animal mesh, $d$ determines the resolution of the texture and each element in $T_i$ is in the range $[0,1]$. 
In our experiments, $f = 4848$ and $d=4$. 
Each $T_i$ is combined into a single matrix, $T \in \mathbb{R}^{fddd3 \times 12}$.
Eigenvectors $E$ are calculated from $T-\bar{T}$, where $\bar{T}$ is the mean of $T$.
Let $f(Y)$ be the function that constrains each $y \in Y$ to be in the range [0,1] and $\beta_{tex}$ be the texture coefficients. A new texture $T'$ can be created as $T' = f(E\beta_{tex} + \bar{T})$. 

After shape and texture of the dog have been generated, the skeleton pose $\theta_{pose}$ of the dog in the form of joint rotations is sampled from the motion dataset of Kearney \etal \cite{kearney_rgbd-dog_2020}. The root of the skeleton is given a random rotation $\theta_{root}$, constrained to result in a mostly upright position. Linear blend skinning is used to apply the pose to the dog mesh, using the skinning weights $W$.

\begin{figure}
    \centering
    \includegraphics[width=\columnwidth]{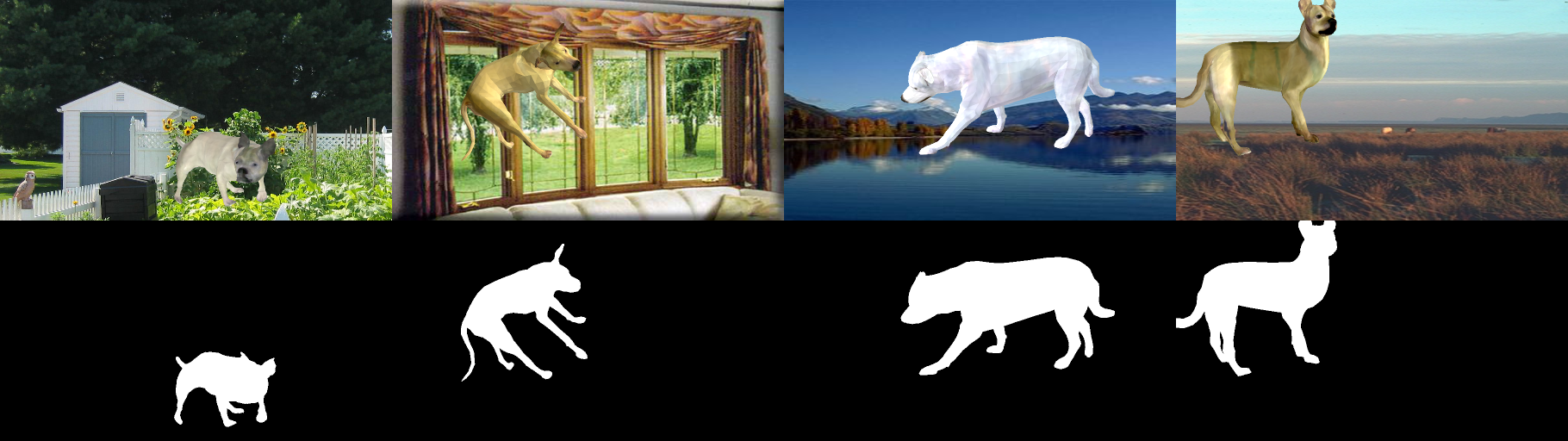}
    \caption{Example RGB images from the synthetic DynaDog+T dataset (top) and their corresponding segmentation masks (bottom).}
    \label{fig: model_img_seg}
\end{figure}

To generate a synthetic dataset with visual qualities similar to a real dataset, we sampled approximately 10,000 images of dogs from the set of approximately 20,000 images described in Section \ref{sec:data}, in order to learn more realistic placement of dogs within the image frame. The 2D joint position information for each image in this dataset was used to record bounding box size and centre positions. As such, we have a distribution from which we can infer the dog’s distance from the camera, that is, a large bounding box with respect to image size indicates the dog is near to the camera, and the inverse is true for a small bounding box.

We focus first on positioning the generated dog in 3D space. The root of the skeleton is fixed for two axes and can move only with respect to distance from the camera, with manually-defined bounds. The distribution of bounding box sizes described
above is scaled to these bounds, and the final root position $d_{root}$ is sampled from the resulting distribution. 
The resulting textured 3D dog mesh is lit with random ambient and directional lighting conditions and rendered as an RGB image and corresponding segmentation map , $256\times455$ pixels in size, via the method of Kato \etal \cite{kato_neural_2017}. This image aspect ratio reflects the sampled real images.

Due to the 3D root constraints, the projected 2D dog will always be positioned in the center of the image (i.e. the dog will be positioned within the bounds of the image). The bounding box of this dog is calculated. We find all instances where the size of the sampled bounding boxes are within $\pm10\%$ of this new bounding box. A Gaussian distribution is created from the corresponding bounding box centres of these instances, and we sample a new centre position $cp$. The projected dog is positioned in the image such that the center of its bounding box aligns with $cp$. In cases where a section of the dog is rendered outside the image bounds, the alignment with $cp$ is restricted to a single axis. For example, if a section of the dog is rendered past the top of the image, the alignment will cause the dog to move potentially left or right in the image, but it cannot move down the image. The generated RGB images were overlaid on background images of select locations from the dataset of Xiao \etal \cite{xiao_sun_2010}, where efforts were made to ensure these images did not already contain a dog. This pipeline is visualised in the supplementary material.

The model $M(\beta_{shape}, \beta_{tex}, \theta_{pose}, \theta_{root}, d_{root}, W)$ generates each 3D dog mesh.
Let $\Gamma$ be the set of rendering parameters, describing the intrinsic and extrinsic properties of the camera and lighting conditions.
$\Pi()$ is the rendering function, producing the RGB image $I_{rgb}$ and segmentation mask $I_{seg}$, $G()$ is the transform applied to re-position the rendered dog on the 2D image, and $I_{bg}$ is the background image sampled from Xiao \etal \cite{xiao_sun_2010}.
The final equation for generating a synthetic image is given in Equation \ref{eq:synthGen}. Examples of our synthetic data can be seen in Figure \ref{fig: model_img_seg} and in the supplementary material.
\begin{equation}
\begin{gathered}
I_{rgb}, I_{seg} = G\Big(\Pi\big( M(\beta_{shape}, \beta_{tex}, \theta_{pose}, \theta_{root}, d_{root}, W), \Gamma\big) \Big) \\
I_{rgb} = \big(I_{bg} \times (1-I_{seg})\big) + I_{rgb}.
\end{gathered}\label{eq:synthGen}
\end{equation}

\subsection{Data for Creating DynaDog+T}
\label{sec:data}
To enable DynaDog+T to generate images with location properties similar to those of real dog images we used videos of dogs we sampled from a dataset of dog videos we recorded ourselves, sourced from the video dataset Youtube 8M \cite{abu-el-haija_youtube-8m:_2016} and videos under public domain  \cite{unsplash_100_nodate, noauthor_20000_nodate, noauthor_dog_nodate, noauthor_free_nodate, noauthor_free_nodate-1}. These videos are processed using DeepLabCut\cite{mathis_markerless_2018, mathis_pretraining_2019} to produce approximately 20,000 images with 43 paired  2D joint coordinates for the dogs in the images from which we sample approximately 10,000 to train DynaDog+T.

\section{Experiments with Synthetic Binary Segmentation Masks}

\begin{table}
\begin{center}
\resizebox{\columnwidth}{!}{
\begin{tabular}{|l||c|c|}
\hline
Dataset & Num. of Images & Num. of Segmentation Maps\\
\hline\hline
Oxford Pets & 4990 & 4990\\
Stanford &  20579  & None\\
COCO & 4562 & 4562\\
OpenImages & None & 1932\\
DynaDog+T & 30000 & 30000 \\
\hline
\end{tabular}
}
\end{center}
\caption{Total Size of Datasets for GAN Experiment}
\label{tab:data}
\end{table}

As noted above we wish to investigate: Is our synthetic data useful for training models when real data is either insufficient or non-existent? To answer this question, we present a simple experiment; the use of a Generative Adversarial Network (GAN)\cite{goodfellow_generative_2014} framework  to refine existing segmentation models. We work on refining existing segmentation models to save time rather than build a model from scratch and use a GAN procedure as opposed to a supervised learning model as there is little paired real canine image and binary segmentation (a.k.a silhouette) map data thus making a traditional supervised learning model impractical. We present three variations of our experiment using three existing segmentation models which we have adapted for binary segmentation for canines (please see supplementary materials for details). For the generator we use the segmentation models DeepLab\cite{chen_deeplab_2017, chen_rethinking_2017}, FCN\cite{long_fully_2014} and Mask-RCNN\cite{he_mask_2018} all of which were pre-trained on the COCO dataset\cite{lin_microsoft_2014} and are available from torchvision\cite{marcel_torchvision_2010}. For our discriminators we use Standard, Patch\cite{isola_image--image_2018} and Pixel\cite{isola_image--image_2018} the architectural details of which are available in the supplementary material. We use the LSGAN\cite{mao_least_2017} loss for stability and  success in other image to image translation tasks\cite{zhu_unpaired_2017}. Our data is composed of three datasets: real images and maps, synthetic images and maps (from our DynaDog+T model) and mixed images and map (both real and DynaDog+T data). 

Between our dataset and GAN combinations, we would have had 27 permutations of experiments. For the sake of practicality we train all generator and discriminator combinations on the real data alone and then take the best performing combinations for each generator and train the same generator/discriminator combination on the mixed and synthetic data, thus running just 15 permutations. Further training details can be found in the supplementary material. 

\textbf{Segmentation Experimentation Data:}  The three main real segmentation datasets we utilised are COCO \cite{lin_microsoft_2014}, Oxford Pets\cite{parkhi_cats_2012} and OpenImages\cite{kuznetsova_open_2018, benenson_large-scale_2019, krasin_openimages_2017} to obtain images and segmentation maps. We converted the Oxford Pets tri-maps into binary segmentation maps. In addition to the paired segmentation datasets we have also included images with no segmentation maps from the Stanford dogs dataset \cite{khosla_novel_2011} to increase the variety of dog images observe by the models. The total number of images and maps for each real dataset and our DynaDog+T synthetic data are shown in Table \ref{tab:data}. The validation set is comprised of images and maps from the Oxford Pet dataset.  We have two test sets, one comprised of Oxford Pet images and maps of single dogs (O) and one made up of the  Oxford Pet test set with additional COCO images/maps which often include more than one dog and/or other objects such as humans prominently in the image (O+C).

Further details can be found in the supplementary material. 

\subsection{Evaluation}
\begin{figure}
\centering
\setlength{\tabcolsep}{0pt}
\renewcommand{\arraystretch}{0}
    \begin{tabular}{ccccc}
         Image & GT & Real & Mixed & Synth \\

 \includegraphics[scale=0.2]{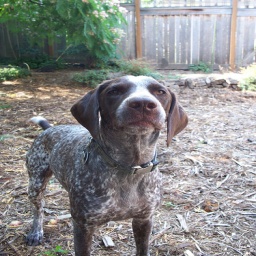}
 & \includegraphics[scale=0.2]{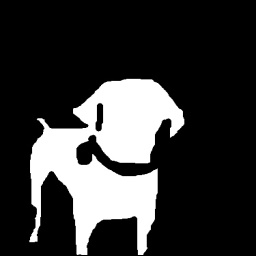}
 & \includegraphics[scale=0.2]{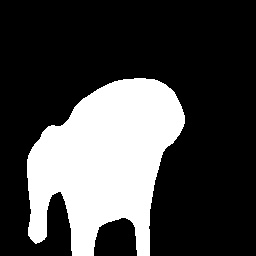}
 &  \includegraphics[scale=0.2]{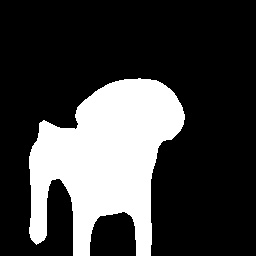}
 & \includegraphics[scale=0.2]{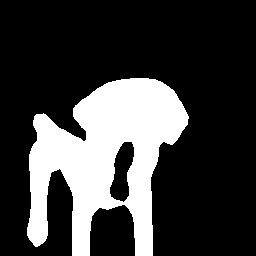} \\

    \includegraphics[scale=0.2]{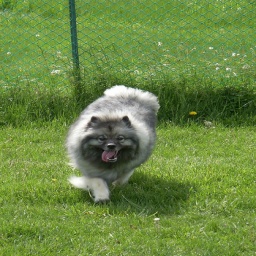}
    &   \includegraphics[scale=0.2]{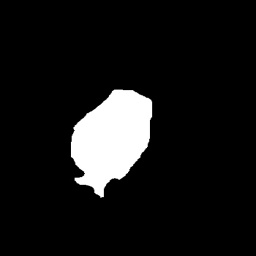}
    &   \includegraphics[scale=0.2]{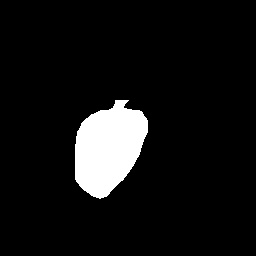}
    &   \includegraphics[scale=0.2]{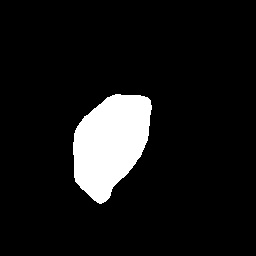}
    &   \includegraphics[scale=0.2]{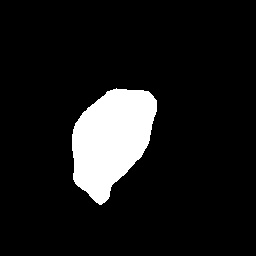}
                       
   \\                    
                       
\includegraphics[scale=0.2]{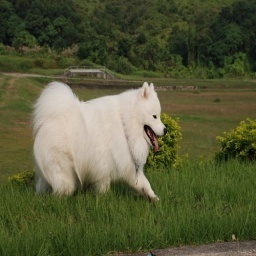}
& \includegraphics[scale=0.2]{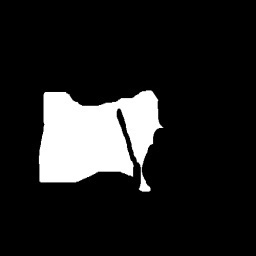}
&   \includegraphics[scale=0.2]{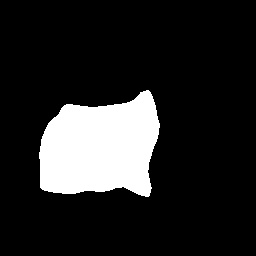}
 &   \includegraphics[scale=0.2]{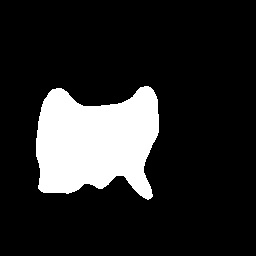}
  &   \includegraphics[scale=0.2]{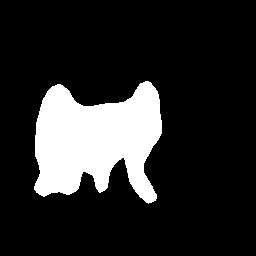}

\\

\includegraphics[scale=0.2]{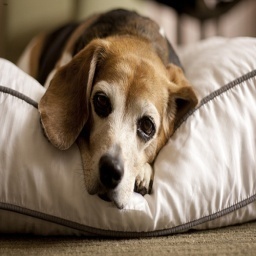}
&   \includegraphics[scale=0.2]{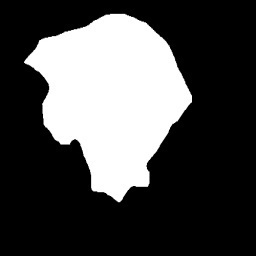}
&   \includegraphics[scale=0.2]{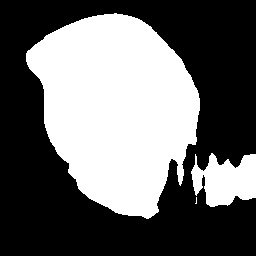}
&   \includegraphics[scale=0.2]{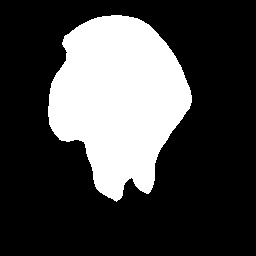}
&   \includegraphics[scale=0.2]{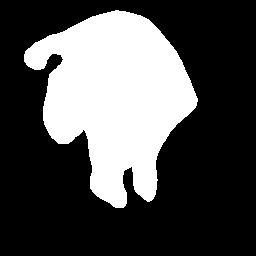}
\\
                        
\includegraphics[scale=0.2]{Images/results/val_image_49.jpg}
&   \includegraphics[scale=0.2]{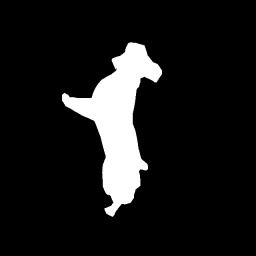}
&   \includegraphics[scale=0.2]{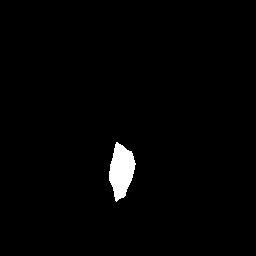}
&   \includegraphics[scale=0.2]{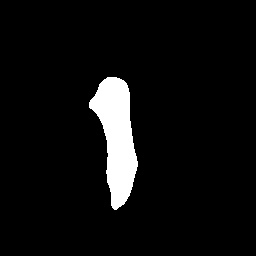}
&   \includegraphics[scale=0.2]{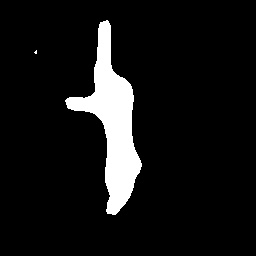}

\\
                        
\includegraphics[scale=0.2]{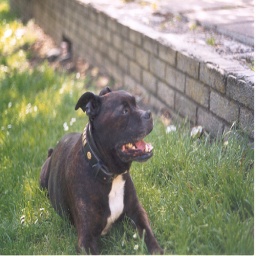}
&   \includegraphics[scale=0.2]{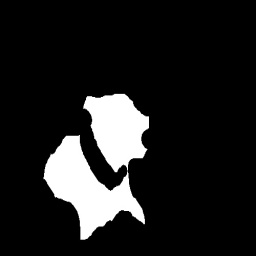}
&   \includegraphics[scale=0.2]{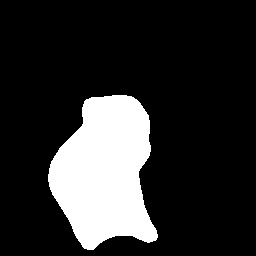}
&   \includegraphics[scale=0.2]{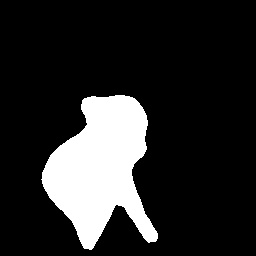} 
&    \includegraphics[scale=0.2]{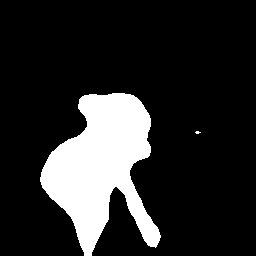} 
\end{tabular}
\vspace{2 mm}
    \caption{Results for DeepLab/Patch GAN. From left to right the columns are: The original image, the ground truth segmentation mask (GT), the predicted masks for the GAN refined using real data (Real), mixed data (Mixed) and  synthetic data (Synth). }
    \label{fig:dp_result}
\end{figure}

We assessed performance based upon binary segmentation maps and thus converted the initial heatmaps produced by the models to binary segmentation maps via the histogram threshold methodology laid out by Ranju \etal \cite{ranju_image_2012} using an initial threshold estimate of 0.7 (which we justify in the supplementary material). For evaluation we have used the Dice/F2 Coefficient, Intersection over Union (IoU) and Pixel Accuracy (Acc). Results are presented in Table \ref{tab:res} while the baseline results, for the unrefined segmentation models, are presented in Table \ref{tab:res_base_0.7_}. In addition results for the best performing GAN are presented in Figure \ref{fig:dp_result}. We find that for DeepLab and FCN, our synthetic data can be used, whether alone or with real data, to refine binary segmentation models to deliver similar or superior performance to those refined on real data alone. This does not hold true for Mask-RCNN where training damages the model performance as can be seen in Tables \ref{tab:res_base_0.7} and \ref{tab:res}.

\begin{table}
\begin{center}
\footnotesize
\resizebox{\columnwidth}{!}{
\begin{tabular}{|l||c|c|c|c|c|c|}
\hline
Segmentor & IoU (O) & F2 (O) & Acc (O) &  IoU (O+C) & F2 (O+C) & Acc (O+C)\\
\hline\hline
DeepLab &  0.3711 &	0.5161&	49.82\%  & 0.3438 & 0.4803 &  51.50\% \\
FCN & 0.3880 &	0.5366 &	56.49\%  & 0.3603 & 0.5010 & 58.55\%\\
Mask RCNN & 0.7143	& 0.8262 &	88.29\%  & 0.7056 & 0.8132 & 88.04\%\\
\hline
\end{tabular}
}
\end{center}
\caption{Results of Baseline Segmentors  for the Oxford test set (O) and the Oxford + COCO test set (O+C) with 0.7 as initial threshold for histogram thresholding.}
\label{tab:res_base_0.7}
\end{table}

\begin{table}
    
    \begin{center}
    \footnotesize
    \resizebox{\columnwidth}{!}{
    \begin{tabular}{|l|l|c|c|c|c|c|c|}
    \hline
         Generator & Discriminator &  IoU (O) & F2 (O) & Acc (O) &  IoU (O+C) & F2 (O+C) & Acc (O+C)\\
         \hline
         \hline
         DeepLab & Standard  & 0.6991	& 0.8112 & 88.34\% & 0.6478 & 0.7554 &	88.86\%\\
    DeepLab & Patch & 0.7471 & 0.8476 & 90.14\% &  0.6903 &	0.7897 &	90.43\% \\
    DeepLab & Pixel &  0.5950 & 0.7283  & 82.83\% & 0.5445 &	0.6719 &	82.01\%	\\

    FCN & Standard  &  0.6191  & 0.7413 & 84.97\% & 0.5684 &	0.6836 &	85.90\%	\\
    FCN & Patch &  0.6390 &	0.7570 &	86.32\% & 0.5788 &	0.6894 & 87.01\%\\
    FCN & Pixel & 0.5590 & 0.7042 &	81.85\% &  0.5130 &	0.6507 & 81.29\%\\

    Mask-RCNN & Standard  & 0.7101 &	0.8229 &	87.99\% & 0.7004	& 0.8090 &	87.83\% \\
    Mask-RCNN & Patch & 0.6783 &	0.7985 &	86.11\%&  0.6679 &	0.7838 &	86.06\%\\
    Mask-RCNN & Pixel &  0.5788	& 0.7176 &	77.99\%&  0.5637 &	0.6979 &	78.38\%\\
         DeepLab (M) & Patch (M) & \cellcolor{red!25}0.7765 & \cellcolor{red!25}0.8678 & \cellcolor{red!25}91.12\% & \cellcolor{red!25}0.7201 &	\cellcolor{red!25}0.8129 & \cellcolor{red!25}91.34\%\\
        FCN (M) & Patch (M) & 0.6661 & 0.7799 & 86.96\% & 0.5146 & 0.6504 &81.21\% \\
        Mask-RCNN (M) & Standard (M) & 0.6760 & 0.7968 & 85.71\% & 0.6672 & 0.7836 & 85.98\% \\
         DeepLab (S) & Patch (S) & 0.6690 & 0.7913 & 86.91\% & 0.6209 & 0.7420 & 87.29\%\\
         FCN (S) & Patch (S) & 0.6236 &	0.7319 & 87.65\% & 0.4854 & 0.6182 & 82.49\%\\
         Mask-RCNN (S) & Standard (S) & 0.6632 & 0.7866 & 85.02\% & 0.6532 & 0.7721 & 85.14\% \\
        \hline
    \end{tabular}
    }
    \end{center}
    \caption{ Results are presented for the Oxford test set (O) and the Oxford + COCO test set (O+C). (M) and (S) are for GANs trained on mixed and synthetic data respectively.  Use of 0.7 as initial threshold for histogram binarisation. Top results are presented in red.}
    
    \label{tab:res}
\end{table}

\section{Conclusion}

We presented our novel parametric canine body model, DynaDog+T. This model enables the generation of synthetic RGB images with associated data points (part segmentation and silhouette maps, 2/3D joint key-points) thereby allowing a work around to animal computer vision problems where comparative real data is sorely lacking. We have also shown that our synthetic data from this model can be used as either a replacement or companion piece to existing real data in training a canine segmentation model. 

However, the generally lower performance of GAN models trained on synthetic data versus real or mixed data appears to indicate that our synthetic data does suffer from a domain gap when compared to real data. This gap is likely due to missing statistical variability present in real data such as fur length, more varied dog shapes or more feasible background image placement. Notably there is also the issue that all of our images contain dogs in  variations of standing poses and there are not sitting or lying  variations on the pose which are otherwise present in the real data. 

In future work we wish to expand upon the shape, pose and texture variability of DynaDog+T's PCA spaces to generate more varied and realistic images with such features as long fur in order to close the domain gap. In addition we would ideally wish to perform a similar comparison for our other synthetic data-points (e.g. 2D joint key-points) but as of the time of writing the amount of ground truth dog data with 2D is not extensive enough to launch a comparable experiment, though with our own data collection methods we hope to alleviate this in the future.


{\small
\bibliographystyle{ieee_fullname}
\bibliography{SynGAN_Seg_CVPRWS.bib}
}

\clearpage

\appendix

\section{Supplementary Material for DynaDog+T: A Parametric Animal Model for Synthetic Canine Image Generation}

This is the companion supplementary material to our main paper. It primarily concerns extra examples of synthetic data generated from DynaDog+T and further details of our segmentation experiment with said synthetic data for binary/silhouette segmentation. We also include further details with regards to model creation and expand our segmentation results with some illustrating graphs.

\section{Further Observations of Animal Vision Problems and DynaDog+T}
\label{obs}

As noted in the main paper, there is a significant amount of work related to computer vision tasks related to human such as pose estimation or 3D mesh reconstruction while there is substantially less for animal related tasks by comparison. There are a number of reasons for this. One of the primary reasons concerns the use of Deep Neural Network (DNN) models for computer vision tasks; they are notoriously data hungry requiring large amounts of data. There is very little existing data (though as noted in the main paper this amount has increased in recent years) for animal vision problems and the creation/collection of such data can be time consuming and expensive. As the main paper stresses, the increasing prominence of synthetic data does allow this issue to be alleviated to a degree. However, even if we do have some framework to generate synthetic data there is another issue that must be considered; the statistical variability of animals in the data. Humans have a fairly general body shape and appearance, albeit with some extremes for the extremely muscular or obese/thin, but nonetheless, there is only so much body shape and phenotype variety among humans. 

In comparison, the intra and inter statistical variability of animals appearance is vast. For example, consider a rhino and an elephant, though they have similar texture, their body shape and other appearance factors are massively different. Comparatively, a german shepard and a boston terrier are both breeds of dog, but possess vastly different physical characteristics. As a result, capturing all the statistical variability present in an animal specie(s) in real data is, practically speaking, virtually impossible. While synthetic data can to some degree alleviate this via the use of PCA spaces for generating new shapes and textures for parametric models, the resulting meshes have textures which 'cling' to the model's mesh (i.e. colour the skin mesh), rather than allowing for features such as hair that flows off the model. A potential solution to such an issue lies in utilising displacement vectors which we add to the shape vector, as has been performed in human works for adding such features as clothes \cite{sengupta_synthetic_2020, alldieck_tex2shape:_2019} and is something we wish to investigate in the future to allow for the generation of more realistic synthetic data in order to close the domain gap with real dog images.

\section{Further Details of DynaDog+T}

In this section, one will find additional details of our DynaDog+T model and auxiliary material that was not included in the main paper. 

\subsection{Bounding Boxes}

\begin{figure}
    \centering
    \includegraphics[width=\linewidth]{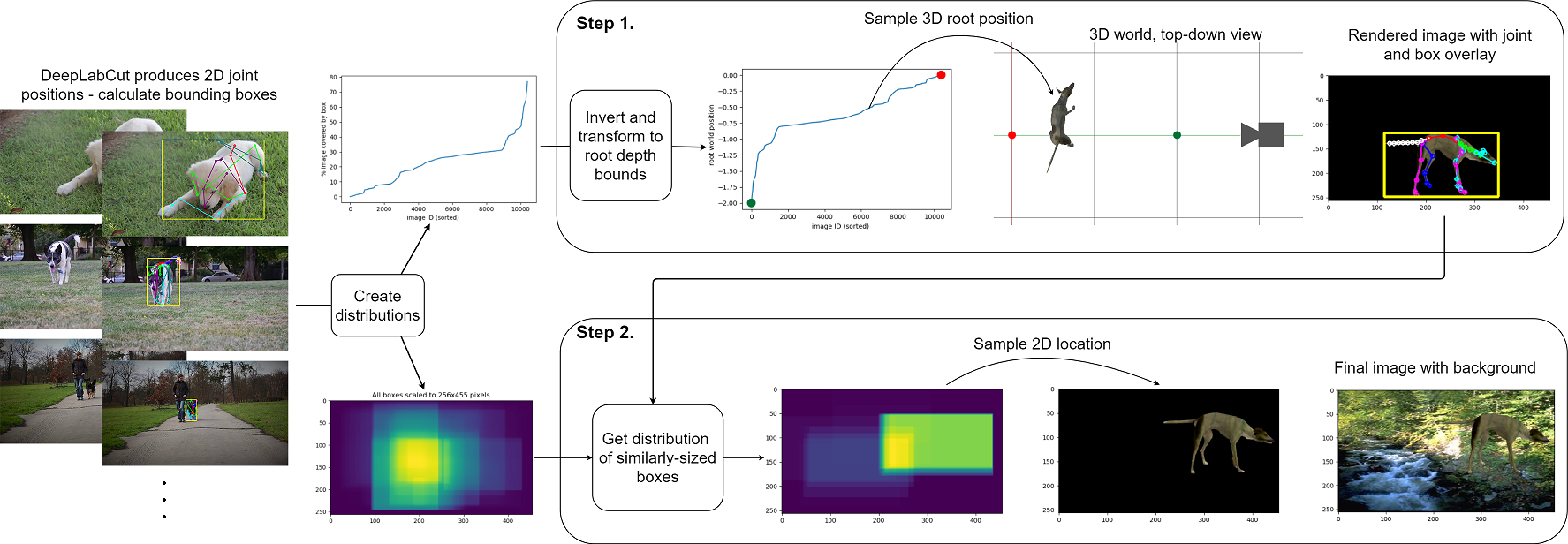}
    \caption{Bounding box information from a selection of real dog images is used as prior information for the positioning of the 3D dog mesh (Step 1) and the location of the rendered dog in the image (Step 2).}
    \label{fig:synthPipeline}
\end{figure}

As described in the main paper, we used dog images with 2D joint locations to build two distributions of bounding box information for our synthetic model; amount of image covered by the box, and the location of the box. The first distribution is used to influence the 3D position of the root of the generated dog skeleton/mesh and the second distribution is used to influence the 2D position of the dog in the rendered image. An illustration of this pipeline is provided in Figure \ref{fig:synthPipeline}

\subsection{Texture}

\begin{figure}
    \centering
    \includegraphics[width=0.9\linewidth]{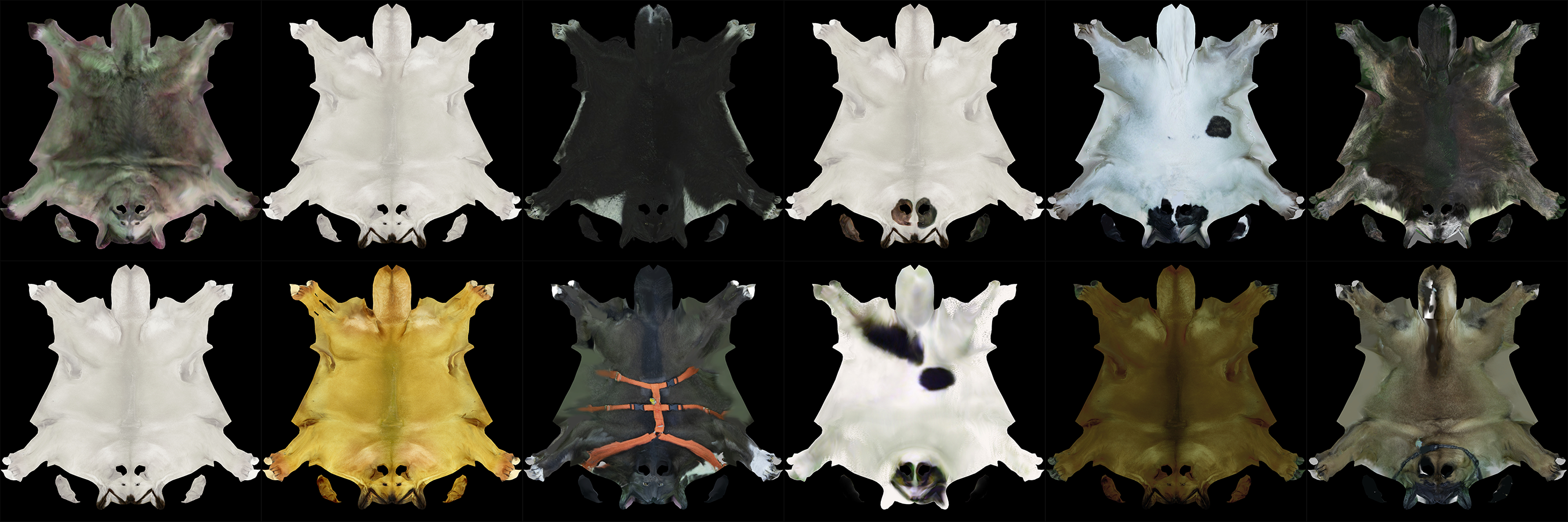}
    \includegraphics[width=0.9\linewidth]{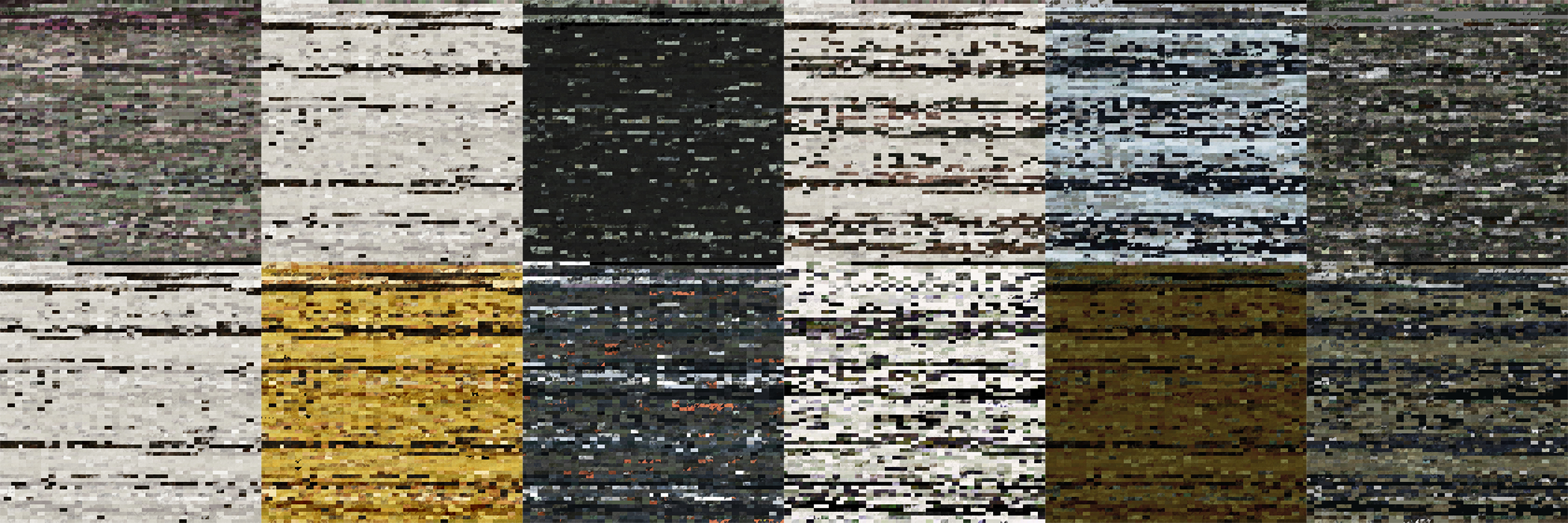}
    \caption{Top: the original dog textures as produced from cleaned photogrammetary scans. Bottom: the textures as produced by Kato \etal\cite{kato_neural_2017}, used to create the PCA model.}
    \label{fig:texture}
\end{figure}

Photogrammetry scans of eleven dogs and a laser scan of a toy dog were used to create a PCA texture model.
These textures were manually refined and processed such that they were globally aligned, and are shown in the top section of Figure \ref{fig:texture}. The PCA model is created from the textures as represented by Kato \etal \cite{kato_neural_2017}, a visual representation of which are shown in the bottom of Figure \ref{fig:texture}.

\subsection{Further Data Instances and Failures}

\begin{figure}
    \centering
    \includegraphics[width=\linewidth]{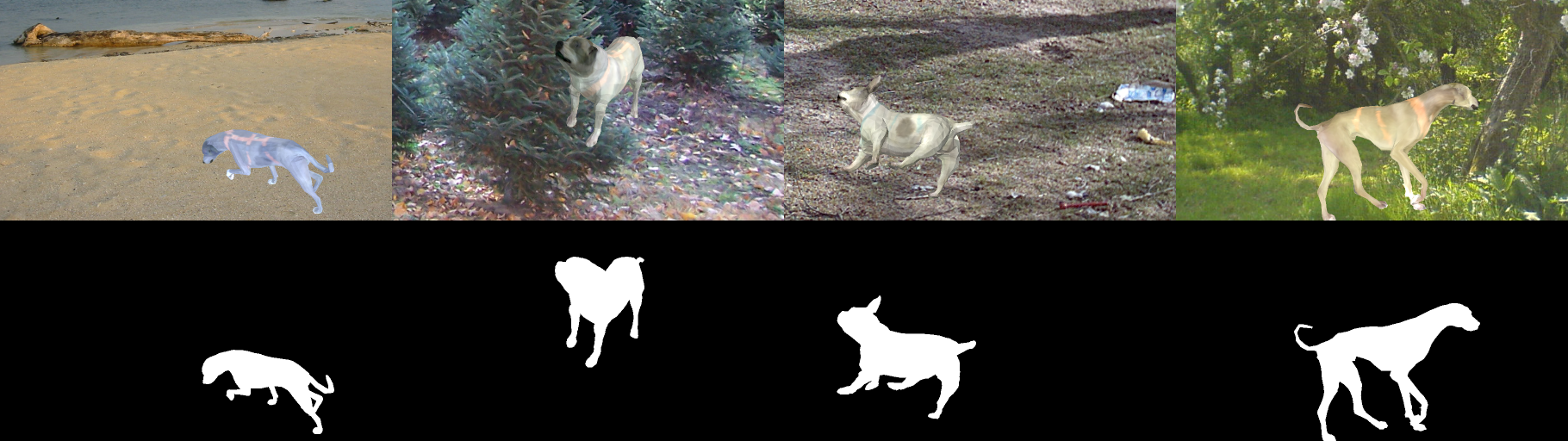}
    \includegraphics[width=\linewidth]{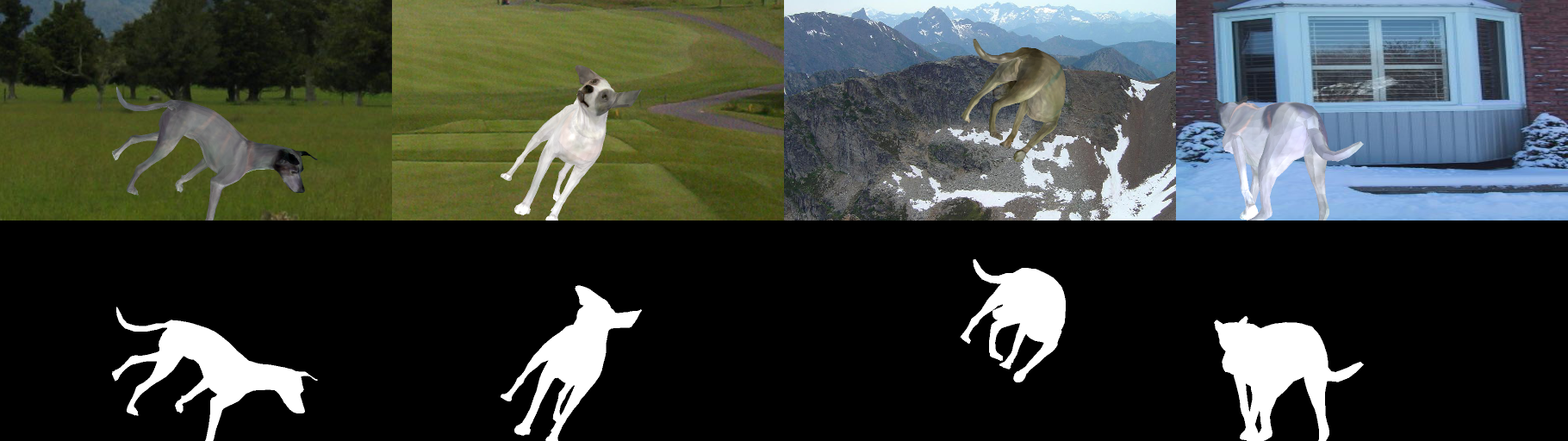}
    \includegraphics[width=\linewidth]{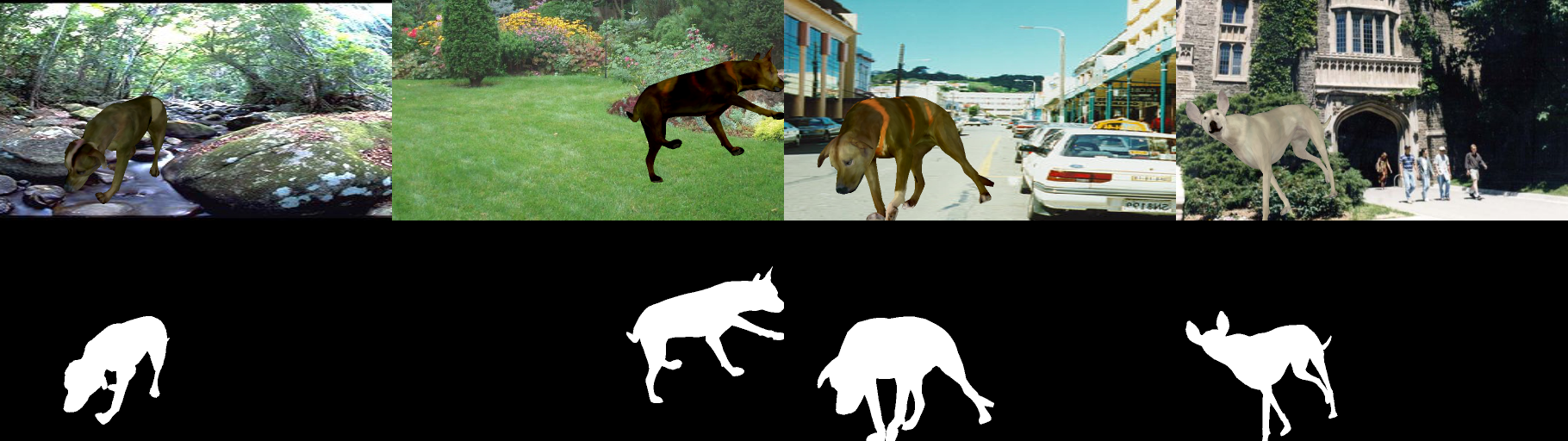}
    \caption{Additional examples from the synthetic dataset}
    \label{fig:synth}
\end{figure}

\begin{figure}
    \centering
    \includegraphics[width=\linewidth]{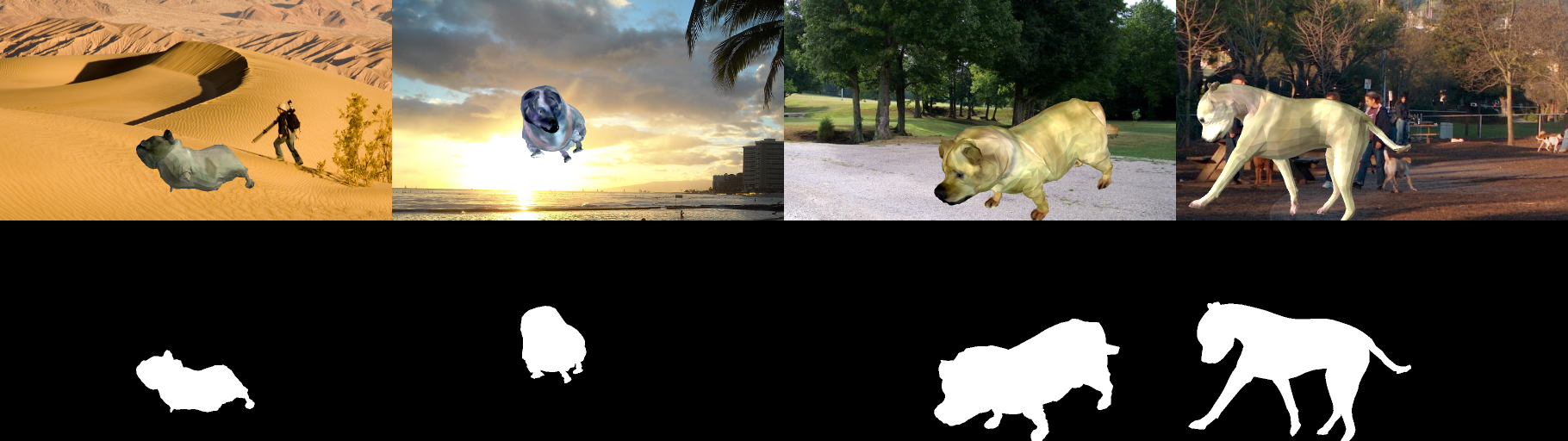}
    \caption{The linear nature of the shape model can produce dogs that have strange proportions. Additionally it is possible for the background image to already contain a dog (right), despite our efforts to prevent this.}
    \label{fig:synthBad}
\end{figure}

Figure \ref{fig:synth} shows further examples of images generated by DynaDog+T with their paired silhouette/binary segmentation maps.  Figure \ref{fig:synthBad} shows some of the issues encountered while generating the dataset: The wide range in variation in dog shape, and the linear nature of the shape model caused the creation of some dogs where the proportions of the dog appears unnatural such as dogs with legs much shorter than their body, or tails unnaturally thin. In addition, although efforts were made to choose background images from Xiao \etal \cite{xiao_sun_2010} that did not already contain a dog, at times such an image was included in the dataset. As a result, the segmentation mask for this image would contain the mask only for the generated dog, and not the pre-existing dog(s). This, understandably can prove problematic for segmentation models like DeepLab.  We should also note that we added a sigmoid layer to each model in order to produce heatmaps with pixel values between 0 and 1 to enable easy thresholding to create binary/silhouette segmentation maps. 

\section{Further Segmentation GAN details and results}

\begin{figure}
    \centering
    \includegraphics[scale = 0.4]{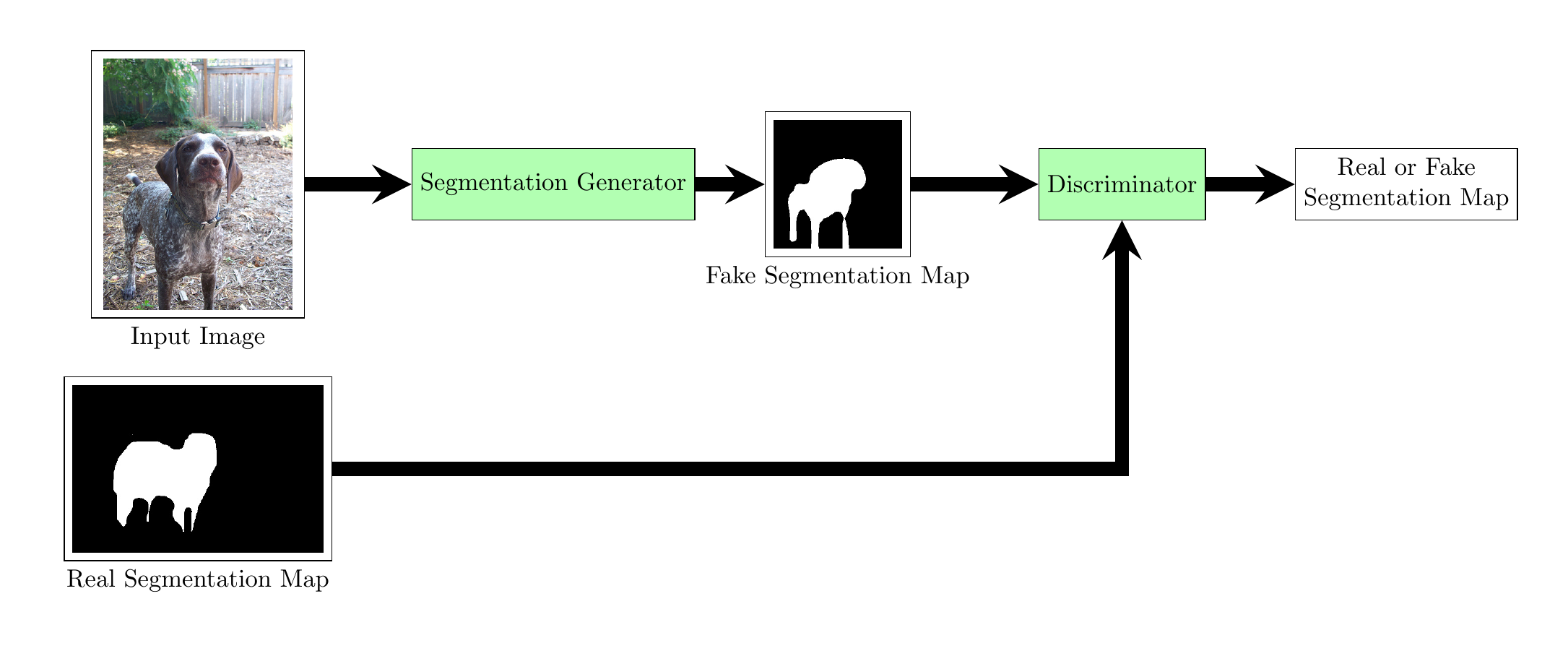}
    \caption{A visual representation of the GAN framework detailed in the main paper}
    \label{fig:framework}
\end{figure}

This section contains further implementation details for our GAN framework expanding upon the section in the main paper. Figure \ref{fig:framework} gives a visualisation of the GAN framework used in the main paper.

It is important to note that during initial code tests we discovered that in some cases  Mask-RCNN\cite{he_mask_2018} could not detect a dog within the image and thus no segmentation map was produced and thus training could not occur as the code would break due to the lack of a predicted map being produced from the generator. Thus we took the naive solution of substituting a blank (i.e. all zero label) map for the segmentation map for any images where a dog was not detected - this is another potential cause of the poor performance of the GANs where Mask-RCNN was the generator, as the training performance would have been damaged as a result. Occasionally for Mask-RCNN multiple dogs are detected in some images and thus multiple segmentation heatmaps are created, we sum across all the non normalised heatmaps produced to create a single heatmap per image (which is then passed to the sigmoid layer), thus creating a semantic segmentation version of Mask-RCNN (as mentioned in the main paper) - which again could be a cause of the poorer performance of Mask-RCNN permutations compared to the baseline. 

In the main paper, we noted that we used three different discriminator architectures: Standard, Patch\cite{isola_image--image_2018} and Pixel\cite{isola_image--image_2018}. Their architectures are given below: 

\begin{itemize}
    \item Standard - Consists of 7 convolutional layers each with a $4 \times 4$ kernel and respective  channels $\{64, 128, 256, 512, 512, 512, 1\}$ with a stride of 2 across all convolutions. Most convolutions are followed by a batch normalisation and Leaky-ReLU paramaterised by 0.2. The exceptions are the first convolution, which lacks a batch normalisation, and the last layer which lacks both normalisation and Leaky-ReLU. Produces a scalar output. 
    \item Patch - Consists of 6 convolution layers with $4 \times 4$ kernel and $\{64, 128, 256, 512, 512, 1\}$ channels with a stride of 2 for all but the last two convolutions. All but the first and last convolutions are followed by batch normalisation and Leaky-ReLU paramatarised by 0.2; the first convolution is also followed by a Leaky-ReLu layer. Produces a $ 14 \times 14$ array where each element corresponds to a  $142 \times 142$ square of the input segmentation map. 
    \item Pixel - Similar to Patch albeit with $\{64, 128, 1\}$ channels with a stride of 1 and no padding. Produces a  $256 \times 256$ array where each element corresponds to a single pixel in the input, containing a probability value for if the pixel in the input comes from a real or fake image. 
\end{itemize}

The objective functions for the networks in our GANs, given in equation \ref{eq:lsgan}, are as defined by Mao \etal \cite{mao_least_2017} where $z$ are the RGB images which we feed into the generator(s) and $x$ is the distribution of the real segmentation maps.

\begin{equation}
\begin{split}
 \min_D L_{LSGAN}(D) &= \frac{1}{2}E_{x\sim p_{data}(x)}[(D(x) - 1)^2] \\ &+ \frac{1}{2}E_{z\sim p_{z}(z)}[(D(G(z)))^2]. \\
 \min_G L_{LSGAN}(G) &= \frac{1}{2}E_{z\sim p_{z}(z)}[(D(G(z)) - 1)^2]. 
 \end{split}
 \label{eq:lsgan}
\end{equation}

\subsection{Implementation}
This project was carried out via PyTorch \cite{paszke_pytorch_2019} using a single NVIDIA GeForce GTX 1070 GPU. We found that, when using a image size of $256 \times 256$ and a batch size of 6, a training time of approximately three to six hours, depending upon network and dataset choices, was sufficient enough to obtain good quality binary segmentation maps, and avoid modal collapse. As mentioned in the main paper, we took the best performing GANs for each generator/discriminator combination trained on real data and trained those same combinations of on the DynaDog+T synthetic data and the mixed dataset.  We used an Adam Optimiser \cite{kingma_adam_2017} for both the generator and discriminator which we update jointly.

\noindent \textbf{Dataset Sizes: } As mentioned in the main paper we used a mix of existing real datasets to build our real silhouette/binary segmentation dataset. Our real training data is composed of 28,455 images with 9,810 binary segmentation maps. Complementing this is the 30,000 images and  binary segmentation maps. As a result our mixed dataset is comprised of 58,455 images and 39,810 maps.  Our validation set is comprised of 748 images and corresponding maps from the Oxford dataset \cite{parkhi_cats_2012} while our Oxford test set (O) comprises 749 images and maps with the Oxford and Coco\cite{lin_microsoft_2014} (O+C) dataset adding an additional 177 images and maps. 

\noindent \textbf{Training on Real data Only: }The generator and discriminator learning rates are given in Table \ref{tab:lr}. We trained all the GAN models for 7500 iterations, as determined by hyper-parameter optimisation. We used the segmentation maps outputted during the training process as an additional qualitative validation of model performance to ensure we were not suffering fro modal collapse.

\noindent \textbf{Training on Mixed data (Real + DynaDog+T):} Due to the results on the validation set we trained for 10,000 iterations before testing, as determined by hyper-parameter optimisation, as this allowed the avoidance of modal collapse while also producing the best results on the validation set. From hyper-parameter optimisation we kept the learning rates from Table \ref{tab:lr} for the generators and used a learning rate of  $5 \times 10^{-7}$ for the discriminators.  

\noindent \textbf{Training on DynaDog+T Only:} From hyper-parameter optimisation, we kept the same generator learning rate as for the real (Table \ref{tab:lr}) and mixed data but took learning rates of $5 \times 10^{-8}$ for the discriminators as we found anything greater lead to modal collapse. We trained for only 5000 iterations due to the results of the validation set on various iterations of the models.

As noted in the main paper, our segmentation models produce heatmaps which we turn into binary segmentation maps via the use of a threshold value. In the main paper, we noted that we chose an initial value of 0.7 for histogram adaptive thresholding\cite{ranju_image_2012}, and  that we used the baseline segmentation models to justify this initial choice. In this supplementary material, we wish to expand upon this justification a little. From Tables \ref{tab:res_base_0.5} and \ref{tab:res_base_0.7_} we can see the results on the Oxford test dataset of the baseline segmentors when we apply an initial threshold of 0.5 and 0.7 respectively for adaptive thresholding. As we can see the difference in threshold has only marginal effect on the value of our evaluation metrics for DeepLab and FCN, with a change of approximately between 0.01 and 0.07 for most metrics. Conversely however, we see significant change for Mask-RCNN which sees massive increases in the value of the metrics, thus justifying our choice of 0.7 as an additional threshold. 

\begin{figure}
    \centering
    \begin{subfigure}[b]{0.3\textwidth}
            \centering
            \includegraphics[width=\linewidth]{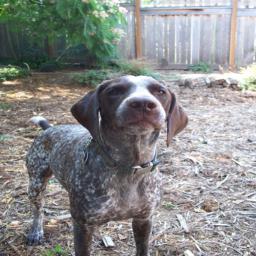}
            \caption{Original Image}
    \end{subfigure}
    \begin{subfigure}[b]{0.3\textwidth}
            \centering
            \includegraphics[width=\linewidth]{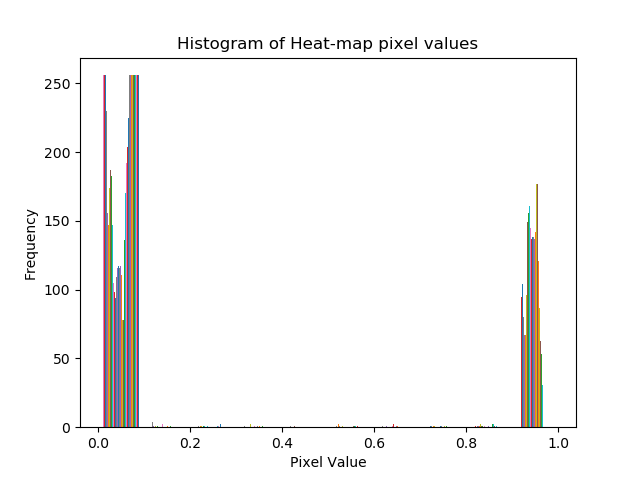}
            \caption{DeepLab Histogram}
            \label{subfig:d}
    \end{subfigure}
    \begin{subfigure}[b]{0.3\textwidth}
            \centering
            \includegraphics[width=\linewidth]{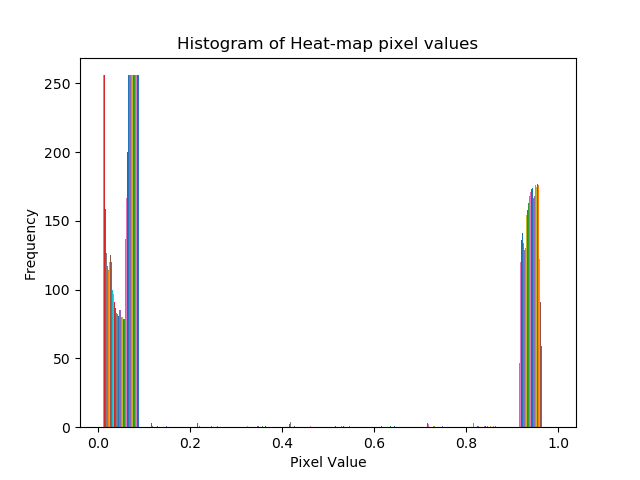}
            \caption{FCN Histogram}
            \label{subfig:f}
    \end{subfigure}

    \begin{subfigure}[b]{0.3\textwidth}
            \centering
            \includegraphics[width=\linewidth]{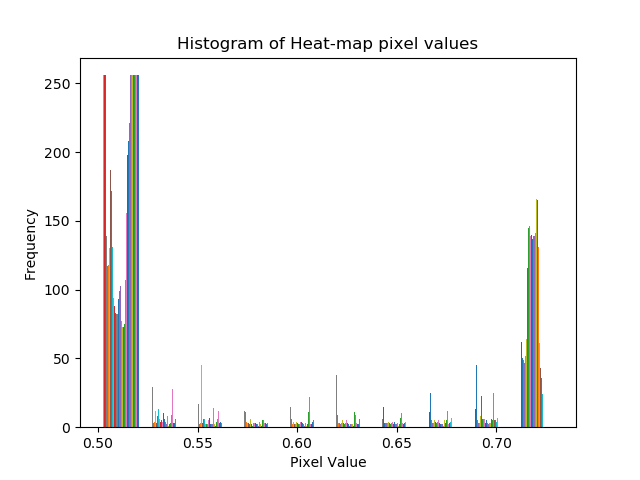}
            \caption{Mask-RCNN Histogram}
            \label{subfig:m}
    \end{subfigure}
    \caption{Histograms and heatmaps for a single image.
    }
    \label{fig:solo_hist}
\end{figure}

The large change in value for Mask-RCNN can be explained by looking at Figure \ref{fig:solo_hist} which displays the histograms of the pixel values for heatmaps produced by refined segmentation models (where the discriminator was Patch) for a single image (whoich is shown).  As we can see, all the histogram are clearly bimodal. Notably however, there is a significant difference between Figures \ref{subfig:d} and \ref{subfig:f} and Figure \ref{subfig:m}; where those peaks are located. For DeepLab and FCN, those peaks are at approximately 0.1 and 0.9 while for Mask-RCNN the peaks are at approximately 0.525 and 0.725. While these histograms are for refined models rather than the baseline, they nonetheless provide a useful illustration as to why the choice in threshold is so important to the results, albeit using only a single instance. 

\begin{table}
\begin{center}
\resizebox{\columnwidth}{!}{
\begin{tabular}{|l||c|c|c|c|c|c|}
\hline
Network & DeepLab & FCN & Mask-RCNN & Standard & Patch & Pixel \\
\hline
Learning Rate & $10^{-7}$ & $10^{-7}$ & $10^{-8}$ & $ 5 \times 10^{-6}$  & $ 5 \times 10^{-6}$ & $ 5 \times 10^{-6} $ \\
\hline
\end{tabular}
}
\end{center}
\caption{Learning Rates for real data}
\label{tab:lr}
\end{table}

\begin{table}
\begin{center}
\footnotesize
\resizebox{\columnwidth}{!}{
\begin{tabular}{|l||c|c|c|}
\hline
Segmentor & IoU & F2  & Acc  \\
\hline\hline
DeepLab &  0.3449 &	0.4870 &	42.24\%  \\
FCN & 0.3866 &	0.5355 &	56.24\%  \\
Mask RCNN & 0.2830	& 0.4170 &	26.27\%\\
\hline
\end{tabular}
}
\end{center}
\caption{Results of Baseline Segmentors for the Oxford test set with 0.5 as initial threshold for histogram thresholding.}
\label{tab:res_base_0.5}
\end{table}

\begin{table}
\begin{center}
\footnotesize
\resizebox{\columnwidth}{!}{
\begin{tabular}{|l||c|c|c|}
\hline
Segmentor & IoU & F2  & Acc  \\
\hline\hline
DeepLab &  0.3711 &	0.5161&	49.82\%  \\
FCN & 0.3880 &	0.5366 &	56.49\%  \\
Mask RCNN & 0.7143	& 0.8262 &	88.29\%\\
\hline
\end{tabular}
}
\end{center}
\caption{Results of Baseline Segmentors for the Oxford test set with 0.7 as initial threshold for histogram thresholding.}
\label{tab:res_base_0.7_}
\end{table}

\subsection{Further Results and Evaluation}

\begin{figure}
\centering
\setlength{\tabcolsep}{0pt}
\renewcommand{\arraystretch}{0}
    \begin{tabular}{ccccc}
         Image & GT & Real & Mixed & Synth \\

 \includegraphics[scale=0.2]{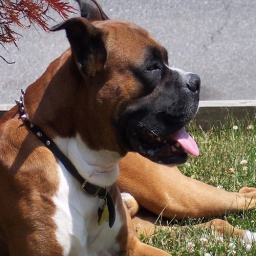}
 & \includegraphics[scale=0.2]{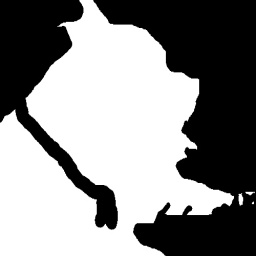}
 & \includegraphics[scale=0.2]{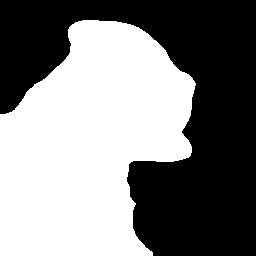}
 &  \includegraphics[scale=0.2]{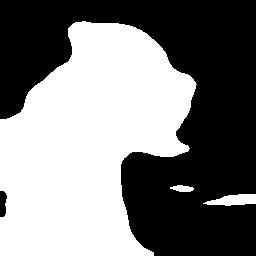}
 & \includegraphics[scale=0.2]{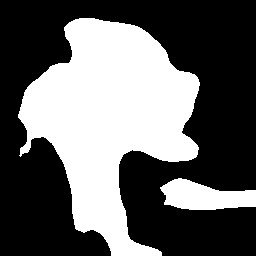} \\

 \includegraphics[scale=0.2]{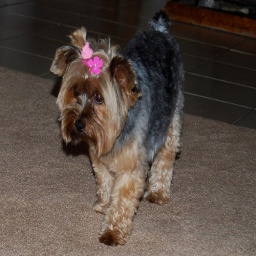}
 & \includegraphics[scale=0.2]{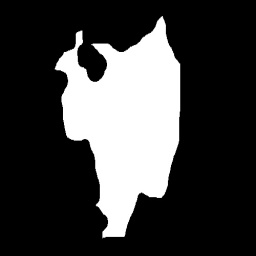}
 & \includegraphics[scale=0.2]{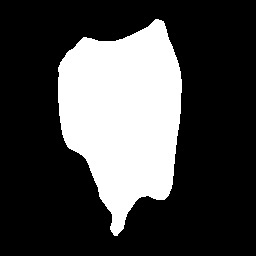}
 &  \includegraphics[scale=0.2]{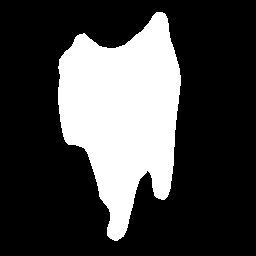}
 & \includegraphics[scale=0.2]{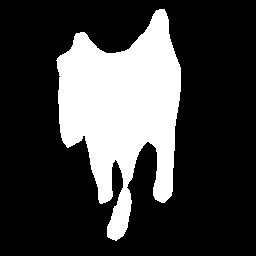} \\
                       
   \\                    
                       
 \includegraphics[scale=0.2]{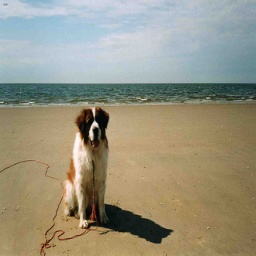}
 & \includegraphics[scale=0.2]{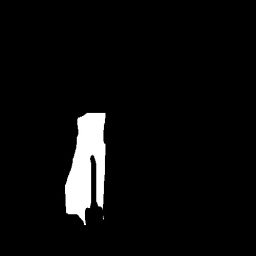}
 & \includegraphics[scale=0.2]{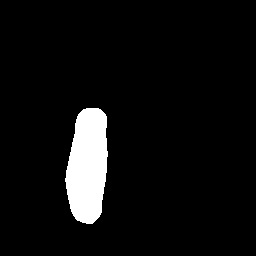}
 &  \includegraphics[scale=0.2]{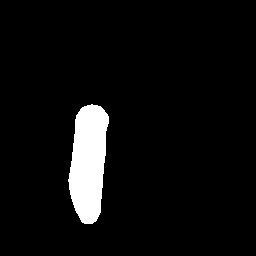}
 & \includegraphics[scale=0.2]{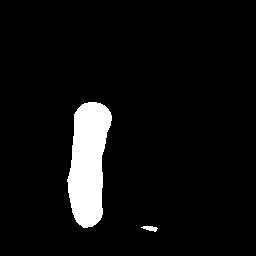}  \\

\\

  \includegraphics[scale=0.2]{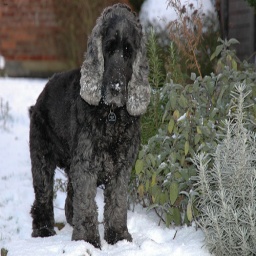}
 & \includegraphics[scale=0.2]{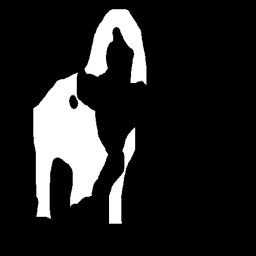}
 & \includegraphics[scale=0.2]{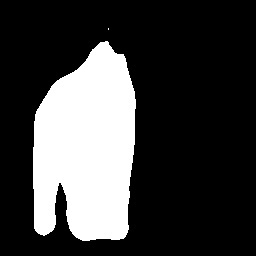}
 &  \includegraphics[scale=0.2]{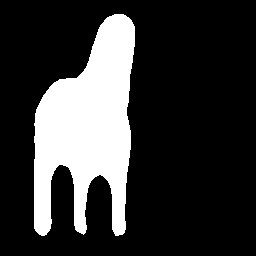}
 & \includegraphics[scale=0.2]{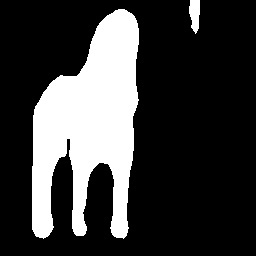}  \\

\\
                        
 \includegraphics[scale=0.2]{Supp_Material/results/4_image.jpg}
 & \includegraphics[scale=0.2]{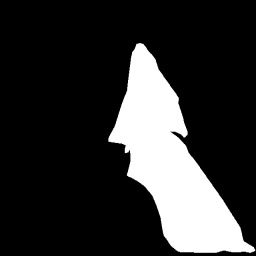}
 & \includegraphics[scale=0.2]{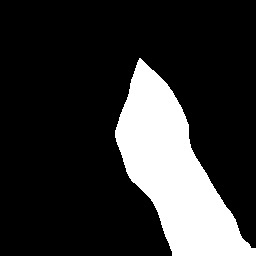}
 &  \includegraphics[scale=0.2]{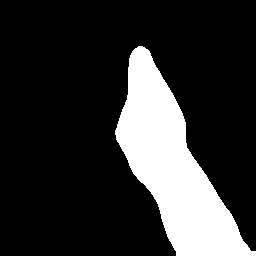}
 & \includegraphics[scale=0.2]{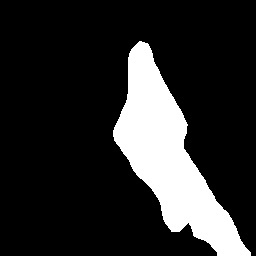}  \\
 
 \\
                        
 \includegraphics[scale=0.2]{Supp_Material/results/5_image.jpg}
 & \includegraphics[scale=0.2]{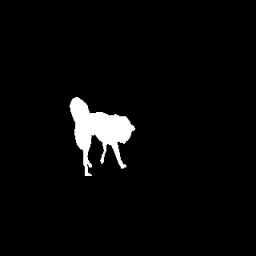}
 & \includegraphics[scale=0.2]{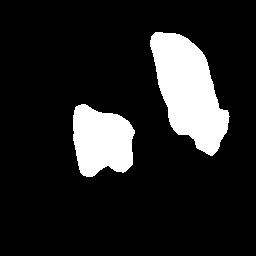}
 &  \includegraphics[scale=0.2]{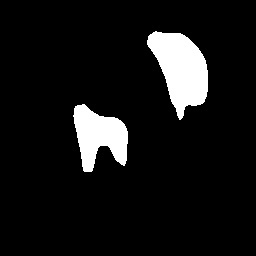}
 & \includegraphics[scale=0.2]{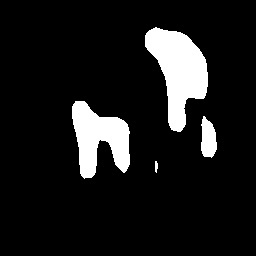}  \\

\end{tabular}
\vspace{2 mm}
    \caption{Further Results for DeepLab/Patch GAN. From left to right the columns are: The original image, the ground truth segmentation mask (GT), the predicted masks for the GAN refined using real data (Real), mixed data (Mixed) and  synthetic data (Synth). }
    \label{fig:res_deeplab_patch}
\end{figure}

\begin{figure}
\centering
\setlength{\tabcolsep}{0pt}
\renewcommand{\arraystretch}{0}
    \begin{tabular}{ccccc}
         Image & GT & Real & Mixed & Synth \\

 \includegraphics[scale=0.2]{Supp_Material/results/0_image.jpg}
 & \includegraphics[scale=0.2]{Supp_Material/results/0_mask.jpg}
 & \includegraphics[scale=0.2]{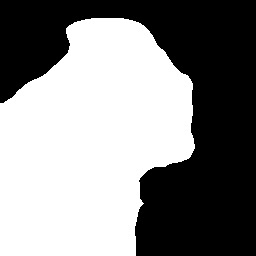}
 &  \includegraphics[scale=0.2]{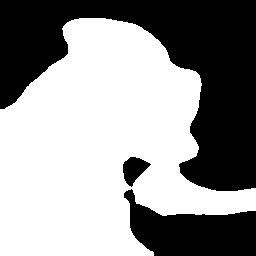}
 & \includegraphics[scale=0.2]{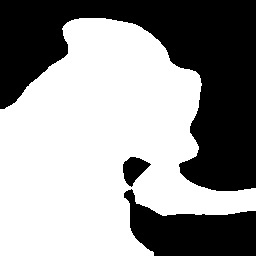} \\

 \includegraphics[scale=0.2]{Supp_Material/results/1_image.jpg}
 & \includegraphics[scale=0.2]{Supp_Material/results/1_mask.jpg}
 & \includegraphics[scale=0.2]{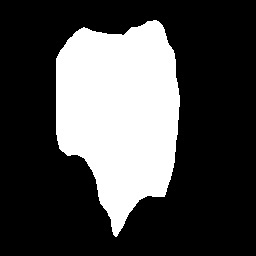}
 &  \includegraphics[scale=0.2]{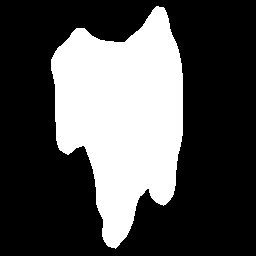}
 & \includegraphics[scale=0.2]{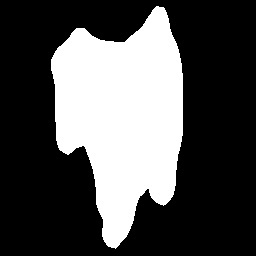} \\
                       
   \\                    
                       
 \includegraphics[scale=0.2]{Supp_Material/results/2_image.jpg}
 & \includegraphics[scale=0.2]{Supp_Material/results/2_mask.jpg}
 & \includegraphics[scale=0.2]{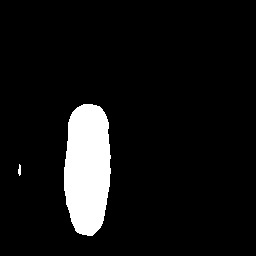}
 &  \includegraphics[scale=0.2]{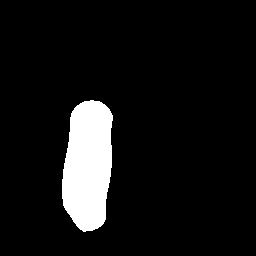}
 & \includegraphics[scale=0.2]{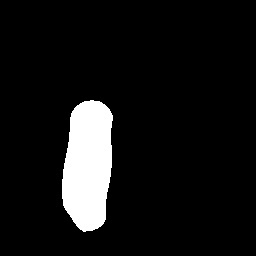}  \\

\\

  \includegraphics[scale=0.2]{Supp_Material/results/3_image.jpg}
 & \includegraphics[scale=0.2]{Supp_Material/results/3_mask.jpg}
 & \includegraphics[scale=0.2]{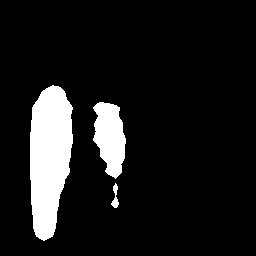}
 &  \includegraphics[scale=0.2]{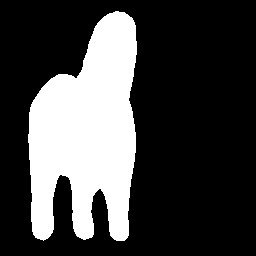}
 & \includegraphics[scale=0.2]{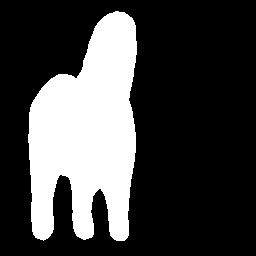}  \\

\\
                        
 \includegraphics[scale=0.2]{Supp_Material/results/4_image.jpg}
 & \includegraphics[scale=0.2]{Supp_Material/results/4_mask.jpg}
 & \includegraphics[scale=0.2]{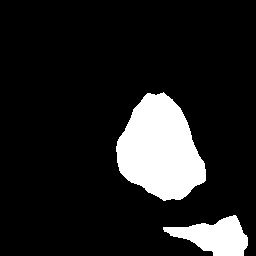}
 &  \includegraphics[scale=0.2]{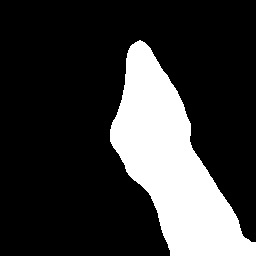}
 & \includegraphics[scale=0.2]{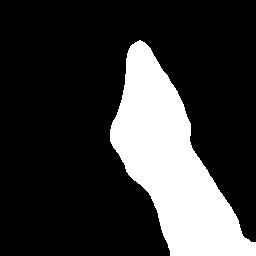}  \\
 
 \\
                        
 \includegraphics[scale=0.2]{Supp_Material/results/5_image.jpg}
 & \includegraphics[scale=0.2]{Supp_Material/results/5_mask.jpg}
 & \includegraphics[scale=0.2]{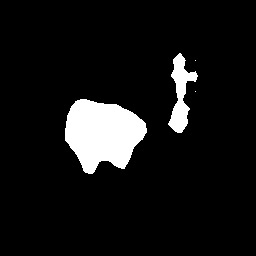}
 &  \includegraphics[scale=0.2]{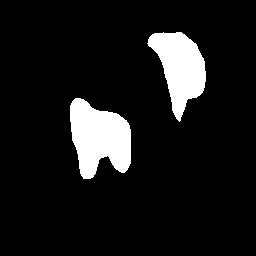}
 & \includegraphics[scale=0.2]{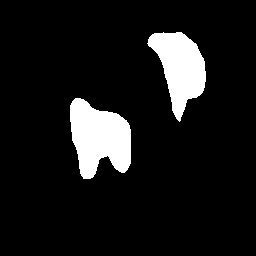}  \\

\end{tabular}
\vspace{2 mm}
    \caption{Further Results for FCN/Patch GAN. From left to right the columns are: The original image, the ground truth segmentation mask (GT), the predicted masks for the GAN refined using real data (Real), mixed data (Mixed) and  synthetic data (Synth). }
    \label{fig:res_fcn_patch}
\end{figure}

\begin{figure}
\centering
\setlength{\tabcolsep}{0pt}
\renewcommand{\arraystretch}{0}
    \begin{tabular}{ccccc}
         Image & GT & Real & Mixed & Synth \\

 \includegraphics[scale=0.2]{Supp_Material/results/0_image.jpg}
 & \includegraphics[scale=0.2]{Supp_Material/results/0_mask.jpg}
 & \includegraphics[scale=0.2]{Supp_Material/results/fp/0_map_r.jpg}
 &  \includegraphics[scale=0.2]{Supp_Material/results/fp/0_map_m.jpg}
 & \includegraphics[scale=0.2]{Supp_Material/results/fp/0_map_s.jpg} \\

 \includegraphics[scale=0.2]{Supp_Material/results/1_image.jpg}
 & \includegraphics[scale=0.2]{Supp_Material/results/1_mask.jpg}
 & \includegraphics[scale=0.2]{Supp_Material/results/fp/1_map_r.jpg}
 &  \includegraphics[scale=0.2]{Supp_Material/results/fp/1_map_m.jpg}
 & \includegraphics[scale=0.2]{Supp_Material/results/fp/1_map_s.jpg} \\
                       
   \\                    
                       
 \includegraphics[scale=0.2]{Supp_Material/results/2_image.jpg}
 & \includegraphics[scale=0.2]{Supp_Material/results/2_mask.jpg}
 & \includegraphics[scale=0.2]{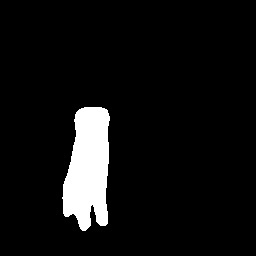}
 &  \includegraphics[scale=0.2]{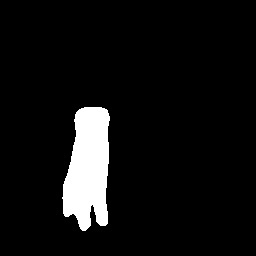}
 & \includegraphics[scale=0.2]{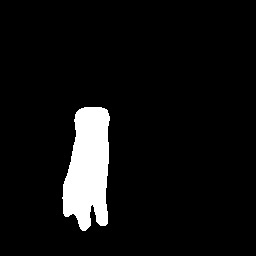}  \\

\\

  \includegraphics[scale=0.2]{Supp_Material/results/3_image.jpg}
 & \includegraphics[scale=0.2]{Supp_Material/results/3_mask.jpg}
 & \includegraphics[scale=0.2]{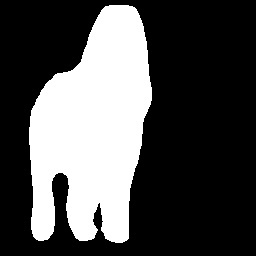}
 &  \includegraphics[scale=0.2]{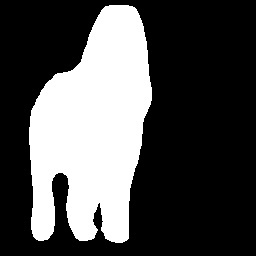}
 & \includegraphics[scale=0.2]{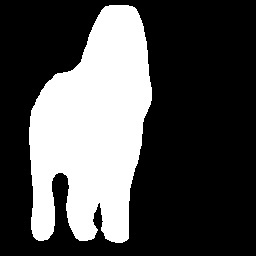}  \\

\\
                        
 \includegraphics[scale=0.2]{Supp_Material/results/4_image.jpg}
 & \includegraphics[scale=0.2]{Supp_Material/results/4_mask.jpg}
 & \includegraphics[scale=0.2]{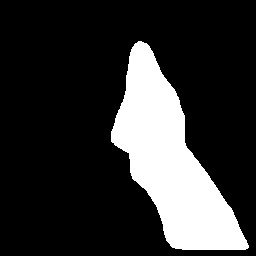}
 &  \includegraphics[scale=0.2]{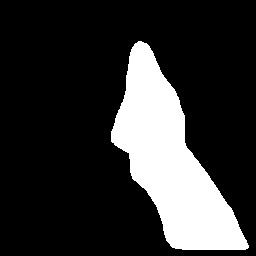}
 & \includegraphics[scale=0.2]{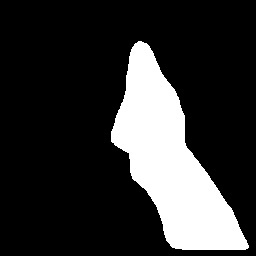}  \\
 
 \\
                        
 \includegraphics[scale=0.2]{Supp_Material/results/5_image.jpg}
 & \includegraphics[scale=0.2]{Supp_Material/results/5_mask.jpg}
 & \includegraphics[scale=0.2]{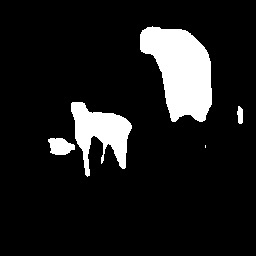}
 &  \includegraphics[scale=0.2]{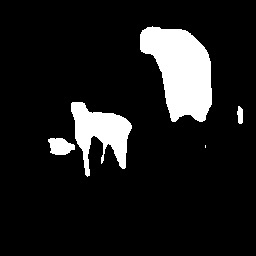}
 & \includegraphics[scale=0.2]{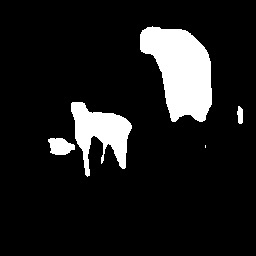}  \\

\end{tabular}
\vspace{2 mm}
    \caption{Further Results for Mask-RCNN/Pixel GAN. From left to right the columns are: The original image, the ground truth segmentation mask (GT), the predicted masks for the GAN refined using real data (Real), mixed data (Mixed) and  synthetic data (Synth). }
    \label{fig:res_mas_s}
\end{figure}

In this section we wish to present some further qualitative results from our refined models. Figures \ref{fig:res_deeplab_patch}, \ref{fig:res_fcn_patch} and \ref{fig:res_mas_s} display results for the DeepLab/Patch, FCN/Patch and Mask-RCNN/Pixel permutations which, as noted in the main paper, were the best performing permutations trained on the real data. We display the image and ground truth maps along with the results for the permutations which were trained on real, mixed and synthetic data. From the Figures, we can see that the results for all the dataset permutations of the GANs do not differ too much, at least in this small subset of images/maps. Interestingly, as we can see from the fourth row our models, that our models are able to learn to fill in silhouettes where the ground truth was otherwise incomplete (we are unsure why this was the case for the ground truth but from an eye test we do believe it to be a very rare event in the dataset). 

That being said, we can also see from the same figures that our models are unable to learn fine silhouette details, such as collars or bows (first and second row), while also and more significantly, in some instances failing to distinguish dogs from other objects as can be seen in the bottom row of the figures, where the human silhouette is also recognised to various degrees. 

\begin{figure}
    \centering
    \includegraphics[width = \columnwidth]{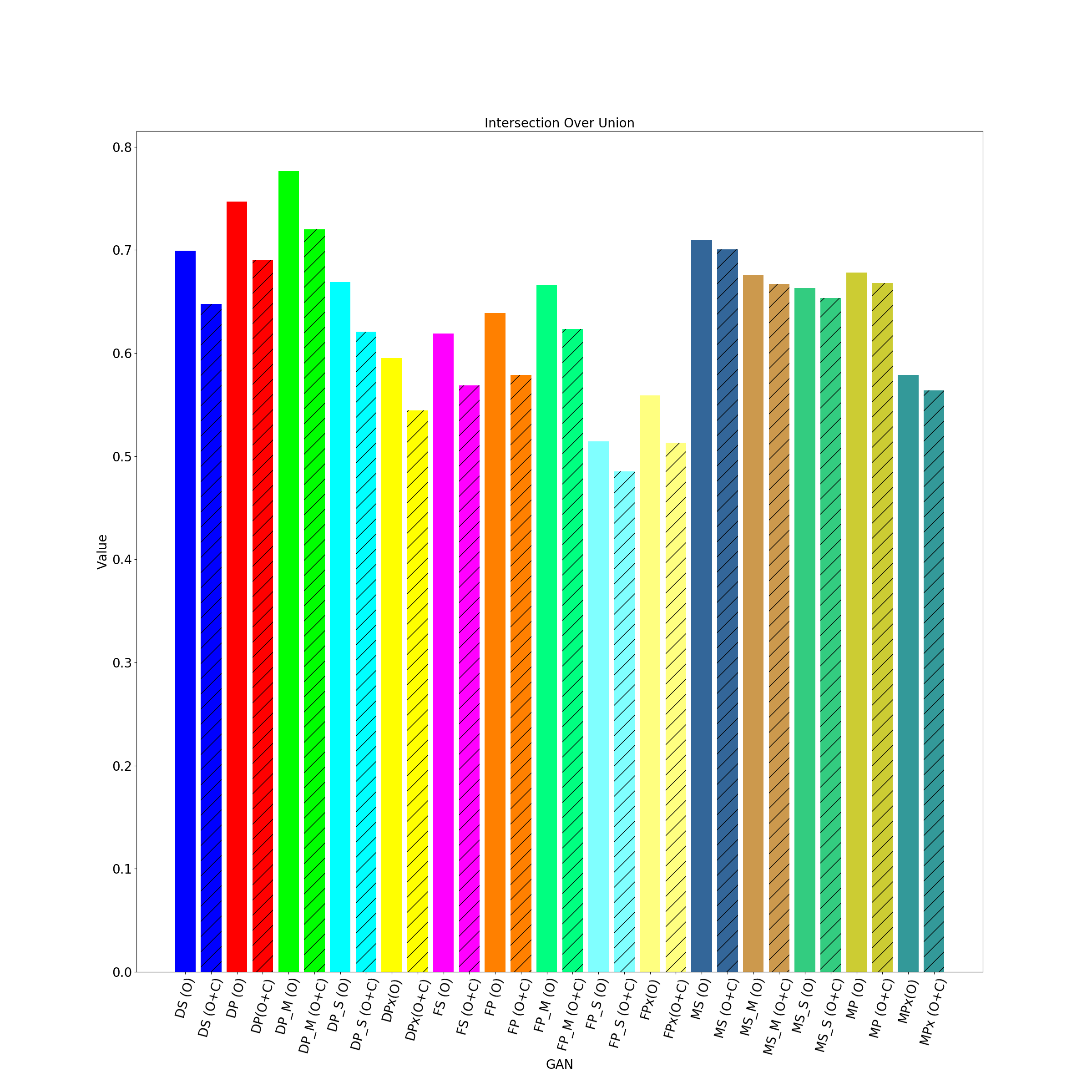}
    \caption{Plot of IOU metric for Oxford (O) and Oxford + COCO (O+C) Test data sets. Models trained on mixed and synthetic Data are denoted by an extension of M and S respectively to the abbreviation.}
    \label{fig:iou}
\end{figure}

\begin{figure}
    \centering
    \includegraphics[width = \columnwidth]{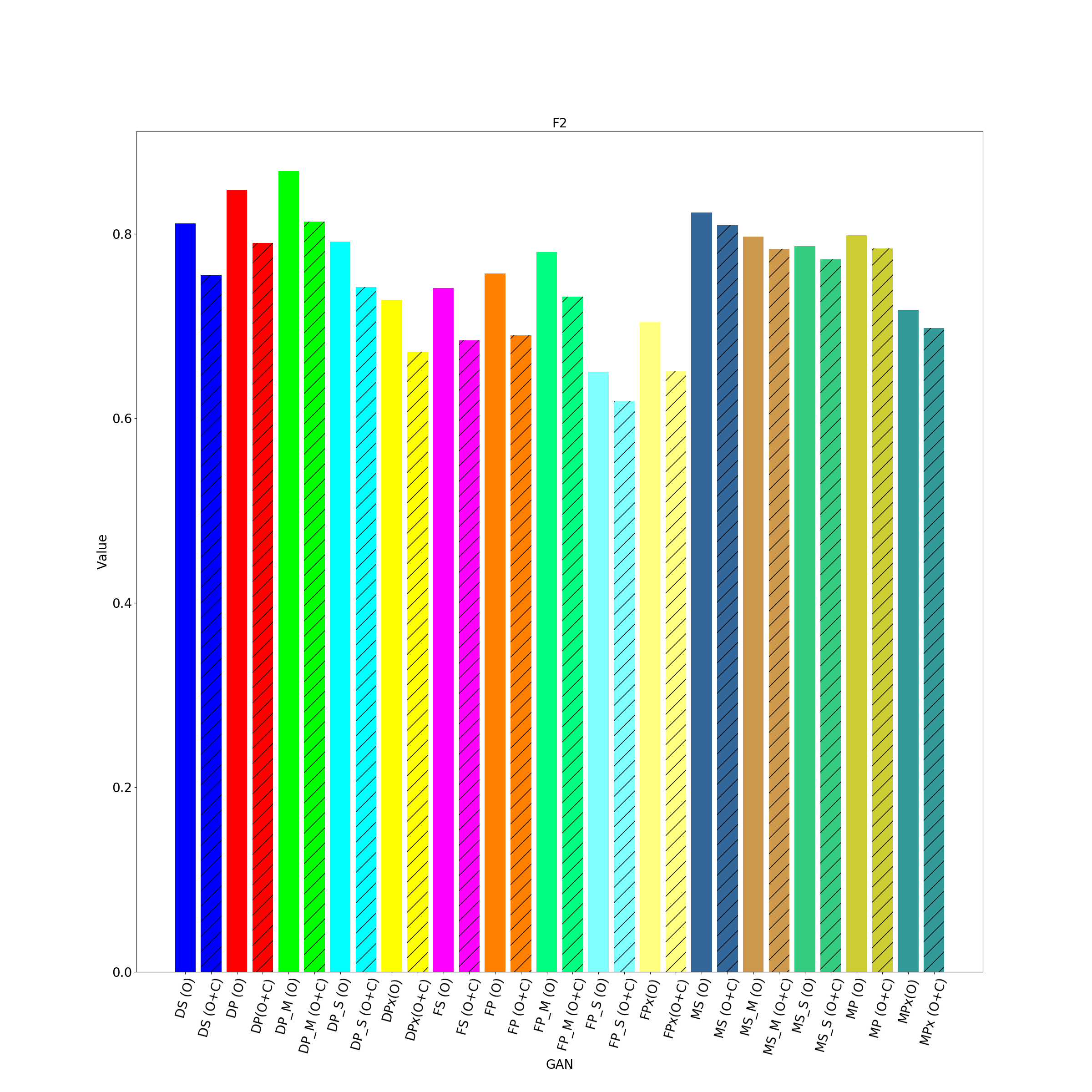}
    \caption{Plot of F2 metric for Oxford (O) and Oxford + COCO (O+C) Test data sets. Models trained on mixed and synthetic Data are denoted by an extension of M and S respectively to the abbreviation.}
    \label{fig:f2}
\end{figure}

\begin{figure}
    \centering
    \includegraphics[width = \columnwidth]{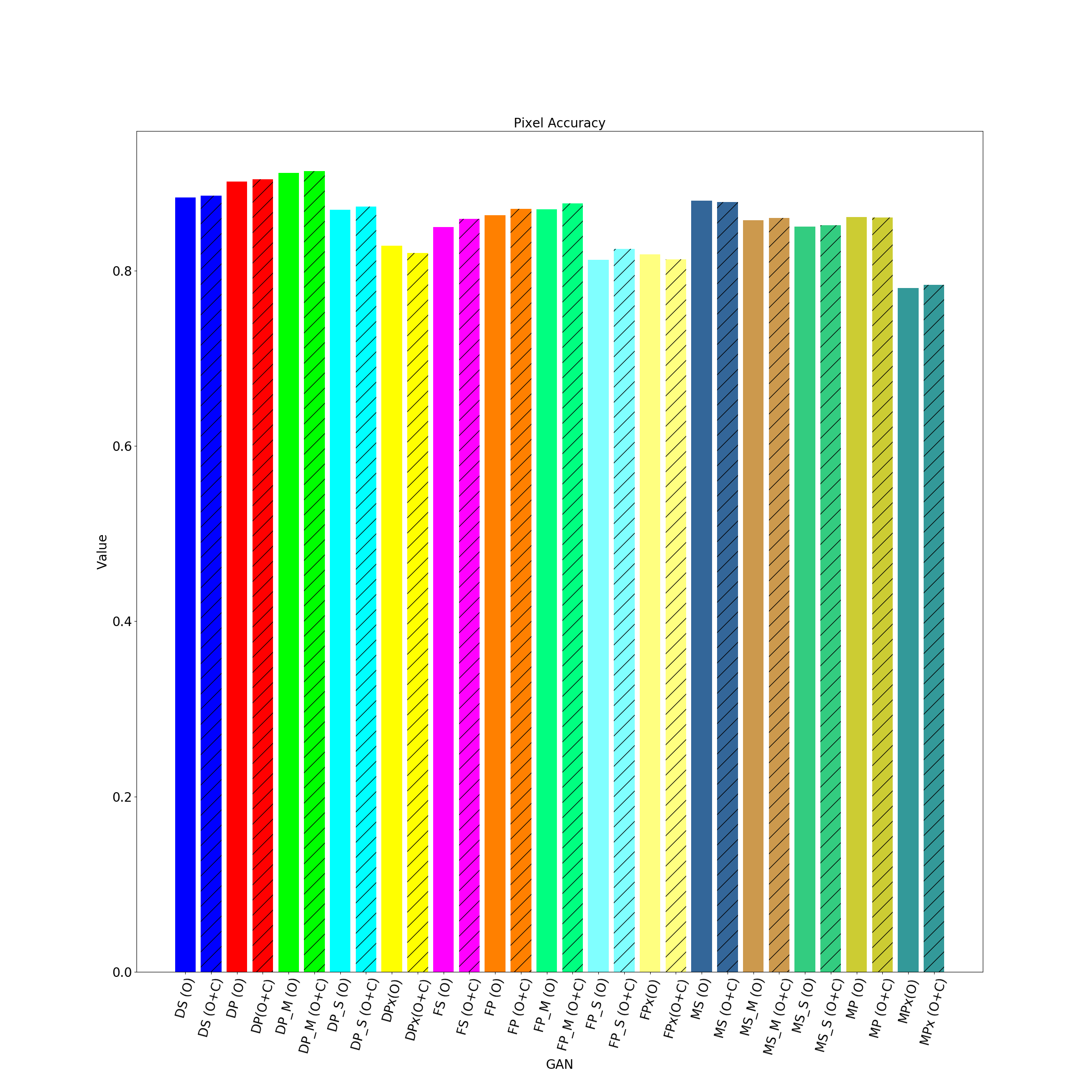}
    \caption{Plot of Pixel Accuracy metric for Oxford (O) and Oxford + COCO (O+C) Test data sets. Models trained on mixed and synthetic Data are denoted by an extension of M and S respectively to the abbreviation.}
    \label{fig:acc}
\end{figure}

We have also provided a graphical summation of the results presented in the table in the main paper through Figures \ref{fig:iou}, \ref{fig:f2} and \ref{fig:acc}. We have plotted all the results for each metric upon the same graph. We have abbreviated the GANs, with the fist letter being that of the Generator (D - DeepLab, F - FCN, M - Mask-RCNN) and the later being those of the discriminators (S - Standard, P - Patch, Px - Pixel). For example DeepLab/Patch is abbreviated to DP. For variants trained upon mixed or synthetic (DynaDpog+T) data, a \texttt{\_M} or \texttt{\_S} is added to the abbreviation. Results on the Oxford and COCO test sets (O+
C) are displayed with diagonal lines on their bars to differentiate them from results for the Oxford (O) test set alone.

As we noted in the main paper, unlike for DeepLab and FCN, our performance for our retrained Mask-RCNN models suffer a noticeable decline compared to the baseline. Our main hypothesis for this is that this is the result of all the changes we had to make to Mask-RCNN in order to get the model to run (these changes were elaborated upon above), which leads to Mask-RCNN learning features that inversely effect performance, notably the use of combining multiple (instance) segmentation into single maps for training could potentially lead to poor results as the resulting maps are unlikely to resemble dogs in some cases.

In creating DynaDog+T and our synthetic data, we are attempting to deal with the unique challenges presented by animal vision tasks. Notably the large amount of variability within certain dog breeds let alone between dog breeds, presents a large barrier for synthetic data to circumvent. In our work here we have attempted to break down, if not all, then at least a part of the barrier. That being said, as noted in the main paper there is still a domain gap with the real data due to the challenge of creating realistic synthetic images with features found in the real animal world. For example, our dog poses are all standing we are missing key features in real dogs such as long fur or collars. Ideally we wish to improve upon our model to further close the domain gap, and allow our data to be used more generally for canine related tasks. As noted in section \ref{obs}, DynaDog+T produces dogs with very specific short haired textures, and we are unable to, as of time of writing, create dogs with long haired textures (such as golden retrievers). To close the domain gap, we will need to increase the realism of our textures along with expanding the statistical variability of our shape and pose  components to enable the generation of images that are closer to realistic images. There is also the issue of poses, as noted in the main paper. Our poses are based upon motion capture data and thus inevitably, DynaDog+T only produces dogs in variations of standing poses as can be seen in Figure \ref{fig:synth}. Understandably this is not representative of the real world variety of poses dogs can adopt as can be seen in Figure \ref{fig:res_deeplab_patch} where we see lying and sitting dogs and as mentioned in the main paper, this is also something we would like to expand upon - most likely through obtaining more varied motion capture data.

\begin{figure}
    \centering
    \begin{subfigure}[b]{0.4\textwidth}
            \centering
            \includegraphics[width=\linewidth]{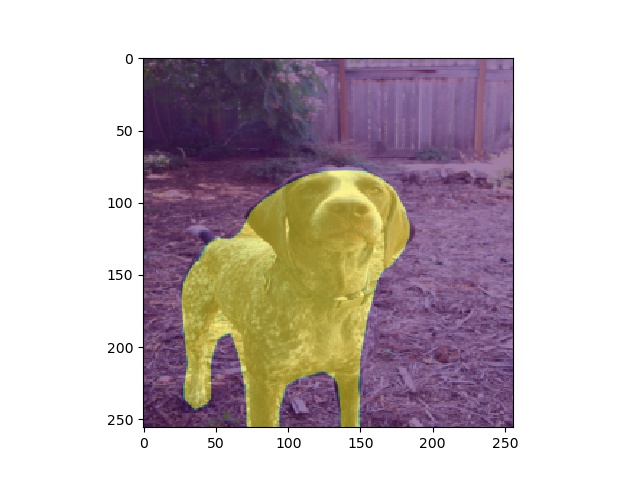}
            \caption{DeepLab Segmentation Heat-Map}
    \end{subfigure}
    \begin{subfigure}[b]{0.4\textwidth}
            \centering
            \includegraphics[width=\linewidth]{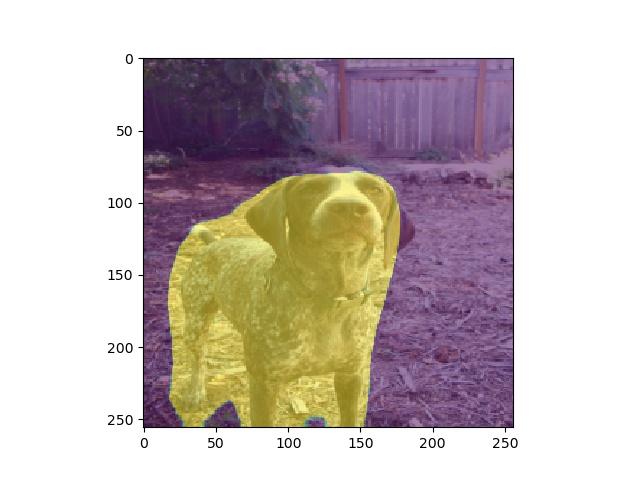}
            \caption{FCN Segmentation Heat-Map}
    \end{subfigure}
    \begin{subfigure}[b]{0.4\textwidth}
            \centering
            \includegraphics[width=\linewidth]{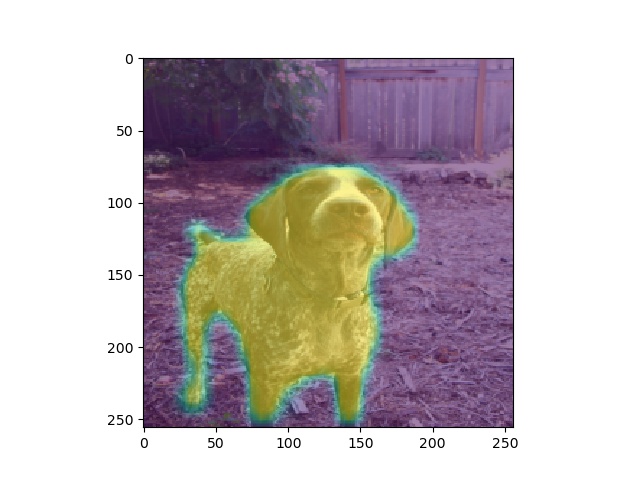}
            \caption{Mask-RCNN Segmentation Heat-Map}
    \end{subfigure}
    \caption{Heatmaps corresponding to the image and histograms in Figure \ref{fig:solo_hist}}
    \label{fig:solo_hm}
\end{figure}

\clearpage

\end{document}